\documentclass[a4paper,fleqn]{cas-dc}

\usepackage[authoryear,longnamesfirst]{natbib}
\usepackage{algorithmic}
\usepackage{algorithm}
\usepackage{float}
\usepackage{orcidlink}

\def\tsc#1{\csdef{#1}{\textsc{\lowercase{#1}}\xspace}}
\tsc{WGM}
\tsc{QE}

\newtheorem{theorem}{Theorem}
\newproof{pot}{Proof of Theorem \ref{thm}}

\begin{document}
\let\WriteBookmarks\relax
\def\floatpagepagefraction{1}
\def\textpagefraction{.001}



\title[mode=title]{Multi-compartment Neuron and Population Encoding Powered Spiking Neural Network for Deep Distributional Reinforcement Learning}  

%

%

\author[1]{Yinqian Sun~\orcidlink{0000-0003-3879-2624}}
\author[1]{Feifei Zhao~\orcidlink{0000-0002-4156-2750}}
\author[1]{Zhuoya Zhao}
\author[1,2,3,4]{Yi Zeng~\orcidlink{0000-0002-9595-9091}}
\cormark[1]
\ead{yi.zeng@ia.ac.cn}
\cortext[1]{Corresponding author}



\affiliation[1]{organization={Brain-inspired Cognitive Intelligence Lab, Institute of Automation, Chinese Academy of Sciences},
            city={Beijing},
            postcode={100190}, 
            country={China}}
\affiliation[2]{organization={Center for Long-term Artificial Intelligence},
	city={Beijing},
	postcode={100190}, 
	country={China}}

\affiliation[3]{organization={University of Chinese Academy of Sciences},
	city={Beijing},
	postcode={100049}, 
	country={China}}
\affiliation[4]{organization={Key Laboratory of Brain Cognition and Brain-inspired Intelligence Technology, Chinese Academy of Sciences},
	city={Shanghai},
	postcode={200031}, 
	country={China}}


%
%
%





\begin{abstract}
Inspired by the brain's information processing using binary spikes, spiking neural networks (SNNs) offer significant reductions in energy consumption and are more adept at incorporating multi-scale biological characteristics. In SNNs, spiking neurons serve as the fundamental information processing units. However, in most models, these neurons are typically simplified, focusing primarily on the leaky integrate-and-fire (LIF) point neuron model while neglecting the structural properties of biological neurons. This simplification hampers the computational and learning capabilities of SNNs.
In this paper, we propose a brain-inspired deep distributional reinforcement learning algorithm based on SNNs, which integrates a bio-inspired multi-compartment neuron (MCN) model with a population coding approach. The proposed MCN model simulates the structure and function of apical dendritic, basal dendritic, and somatic compartments, achieving computational power comparable to that of biological neurons. Additionally, we introduce an implicit fractional embedding method based on population coding of spiking neurons. We evaluated our model on Atari games, and the experimental results demonstrate that it surpasses the vanilla FQF model, which utilizes traditional artificial neural networks (ANNs), as well as the Spiking-FQF models that are based on ANN-to-SNN conversion methods. Ablation studies further reveal that the proposed multi-compartment neuron model and the quantile fraction implicit population spike representation significantly enhance the performance of MCS-FQF while also reducing power consumption.
\end{abstract}



\begin{keywords}
 Machine Learning \sep Brain-inspired Intelligence\sep Spiking Neural Networks\sep Multi-compartment Neuron Model\sep Distributional Reinforcement Learning
\end{keywords}

\maketitle

\section{Introduction}

Spiking neural networks, which emulate the dynamic behavior of biological neurons, offer substantial advantages in terms of energy efficiency and robustness. Recently, neuromorphic computing has successfully applied SNNs across a wide range of domains, including image classification~\citep{vaila2021deep,lee2020enabling,li2022efficient}, object detection~\citep{kim2020spiking,luo2021siamsnn}, speech recognition~\citep{dominguez2018deep, ponghiran2022spiking,wang2024restoring}, decision-making~\citep{tan2021strategy,zhao2018brain,sun2022solving}, and robotic control~\citep{balachandar2020spiking,tang2020reinforcement}. SNNs have demonstrated performance comparable to that of traditional artificial neural networks (ANNs)~\citep{zeng2022braincog}. Moreover, neuromorphic hardware~\citep{merolla2014million,furber2014spinnaker,davies2018loihi} enables the deployment and training of SNNs with low power consumption, while maintaining biological plausibility.

Neurobiological research underscores the critical role of dendrites in integrating synaptic inputs, thereby significantly enhancing the information processing capabilities of neurons~\citep{kampa2006calcium,smith2013dendritic,gidon2020dendritic}. Neurons in the mammalian forebrain exhibit relatively distinct computational compartments, and the  computing function of the individual pyramidal neuron can be mapped to an abstracted two-layer neural network~\citep{poirazi2003pyramidal}. Previous studies predominantly employ multi-compartment neuron (MCN) models for tasks involving simple stimulus classification and image recognition, such as MNIST images~\citep{urbanczik2014learning,sacramento2018dendritic} and neuron spiking sequences~\citep{zhang2024tc,li2023learning}. These models can only handle simple features and are not capable of more complex decision-making tasks, such as playing video games. Besides, typical features of existing MCN computational models are that the neurons have distinct dendritic compartments receiving signals from a common source, which is oversimplified compared to the MCN in the human brain.
In this work, inspired by the structure of pyramidal neurons in the human cortex, we propose a multi-compartment neuron model that includes apical dendrite, basal dendrite, and soma computational compartments. For decision-making task,
we assign decision-making signals from distinct sources to the apical and basal dendrites, with the somatic potential integrating inputs from these compartments. When the somatic potential exceeds a defined threshold, spike outputs are generated~\citep{polsky2004computational,makara2013variable}.

During information processing in the human brain, multi-compartment neurons precisely represent continuous real-valued information in a population encoding manner~\citep{SANGER2003238,PANZERI2015162}. Dynamic population encoding allows the brain to more effectively handle working memory tasks~\citep{meyers2018dynamic}. Inspired by this neural population encoding mechanism, we propose a spiking fractional value population encoding method. This method encodes the fraction within the spike information space, using the resulting population spikes as input signals to the apical dendrite of the MCN neuron, serving as one of the decision-making sources.

Recent research has employed spiking neural networks to address reinforcement learning (RL) problems, leading to the development of models such as spiking DQN~\citep{liu2021human,chen2022deep}. However, these approaches generally rely on simple deep SNN architectures similar to those in vanilla DQN, presenting significant challenges when tackling more complex decision-making tasks, such as those found in distributional reinforcement learning. Distributional reinforcement learning, which estimates the value distribution of returns, more closely mirrors the decision-making processes in the human brain~\citep{lowet2020distributional}. In addition, there are still limitations in point neuron information integration and population neuron information processing for SNNs.

In this work, we introduce biologically plausible SNNs to the domain of deep distributional reinforcement learning for the first time, presenting a multi-compartment spiking fully parameterized quantile function network (MCS-FQF) model. This model combines a multi-compartment neuron framework with a precise population encoding strategy, offering a significant advancement in the capability of SNNs to support complex decision-making tasks.

\begin{figure}[!htb]
	\centering
	\includegraphics[width=1.0\linewidth]{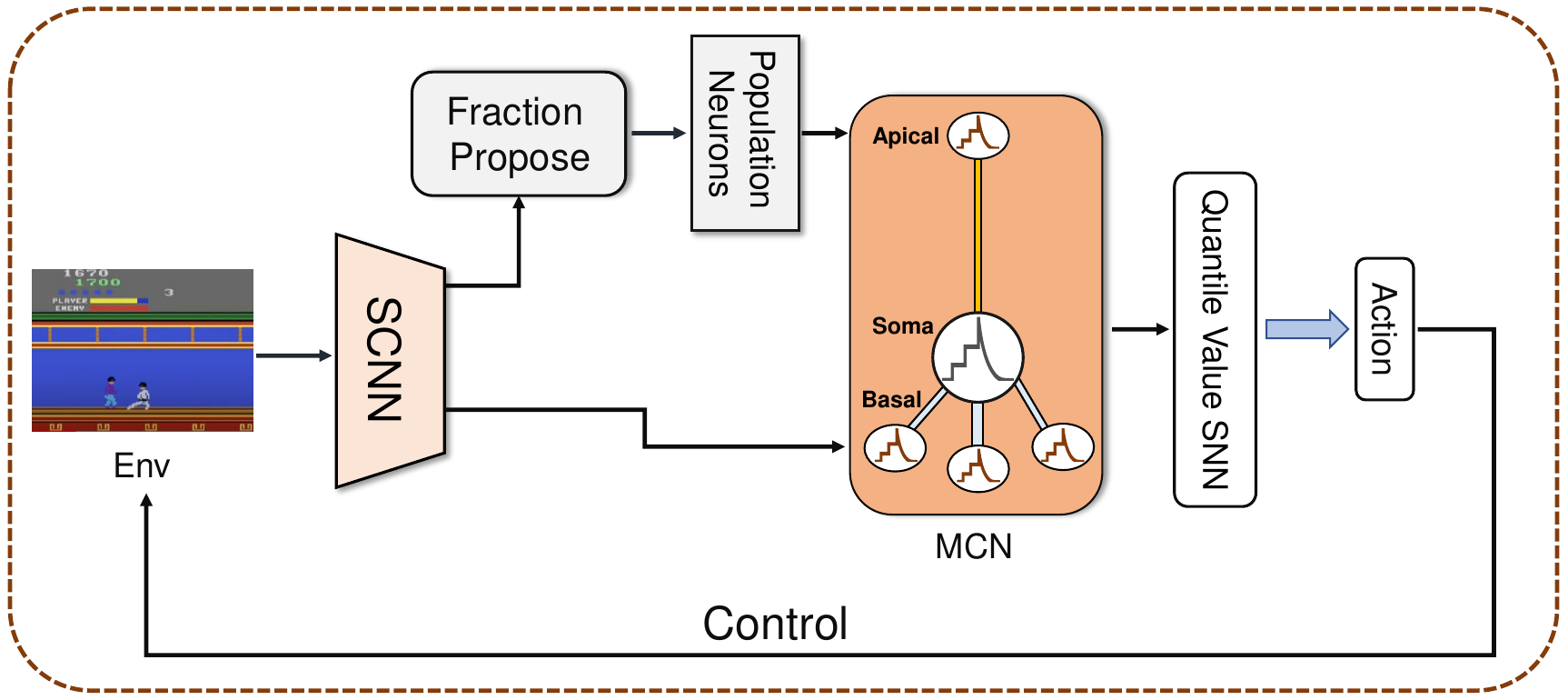} 
	\caption{The overall architecture of MCS-FQF model.}
	\label{Fig:framework}
\end{figure}

As illustrated in Figure~\ref{Fig:framework}, the overall architecture of the proposed MCS-FQF model begins with a spiking convolutional network (SCNN), which encodes game video observation states into spike trains. We then utilize a spiking neuron population encoding method to generate implicit spiking representations of quantile fractions from these state embeddings. The multi-compartment neuron model integrates both state and quantile fraction information: the basal dendrite module receives spike inputs corresponding to the state embeddings, while the apical dendrite module processes the spiking population embeddings of the quantile fractions. The somatic module integrates the information from both basal and apical dendritic potentials. The resulting spike outputs from the multi-compartment neuron represent a combined encoding of state and quantile fraction information, which is further processed by a fully connected SNN to generate the quantile values.

Although several training methods exist for multi-compartment neural networks~\citep{urbanczik2014learning,zhang2024tc}, they are typically restricted to shallow networks and do not scale well to deep structures in complex tasks, such as those encountered in deep reinforcement learning (DRL) environments. In this work, we propose a direct training algorithm for the multi-compartment neuron (MCN) model, employing a surrogate function to approximate the MCN's spike gradient and deriving a spatio-temporal backpropagation (STBP) method~\citep{zhang2021rectified} for optimizing the MCN weights. By training the MCS-FQF model using an end-to-end gradient descent algorithm, we achieve significant performance improvements in the Atari environment. We summarize the main contribution of this paper as follow:

\begin{itemize}
	\item We present a brain-inspired MCS-FQF model to explores the application of SNNs on complex distributional reinforcement learning tasks.
	\item We provide a thorough analysis on various the functions of different structure compartment of biological neurons and build the multi-compartment neuron model to extend the computing capability of a single spiking neuron.
	
	\item The proposed MCS-FQF model takes into account a biologically plausible population encoding method that enables the spiking neural networks to precisely represent the continuous fraction values. And the experiment shows the brain-inspired population encoding method is more efficient than cosine embedding in traditional FQF model for representing the selective scalar fraction values in spiking neural networks.
	
	\item Extensive experiments demonstrate that the proposed MCS-FQF model achieves better performance than the vanilla FQF model and ANN-SNN conversion based Spiking-FQF model on Atari game experiments. Moreover, the ablation study results show that the population encoding and MCN model has irreplaceable contributions in boosting the capability of spike neurons in information representation and integration.
\end{itemize}

\section{Related work}

The LIF ~\citep{gerstner2002spiking} model takes neurons as point-like structures similar to the traditional artificial neuron model, which only uses the somatic membrane potential to process the input signal. Some other spiking neuron models, such as the Izhikevich neuron~\citep{izhikevich2004model} and Hodgkin-Huxley neuron~\citep{hodgkin1952measurement}, have more dynamic characteristics than the LIF model but still do not consider the neurons’ structure characteristics. In addition, the spatiotemporal information processing advantages of the spike neuron model~\citep{shan2024advancing} also need to be further explored. The multi-compartmental neuron model captures the morphology features of biological neurons, employing dynamic equations to model dendritic information processing, membrane potential propagation, and somatic spike generation processes. The dendritic prediction method~\citep{urbanczik2014learning,sacramento2018dendritic} provides a two-compartment neuron model consisting of soma and dendrite modules. It optimizes the dendritic synapse weights to minimize the inconsistency between the dendritic potential and somatic firing rate. Richards et.al~\citep{richards2019deep} proposed a multi-compartment neuron model, consisting of apical dendrites, basal dendrites, and soma, is used to classify the MNIST images. Lansdell et.al~\citep{lansdell2019learning,capone2022burst} proposed apical dendrite target-based learning, the apical dendrite receive the plasticity signal, and neurons' weights can be modified locally. 

The multi-compartment neuron models implemented on neuromorphic chips can simulate the dynamic processes of the brain's neural circuits with extremely low power consumption~\citep{Shrestha2021In,kopsick2022robust}. Yang et.al ~\citep{yang2019scalable} presents a hardware-efficient and scalable strategy for real-time implementation of large-scale biologically meaningful neural networks with multi-compartment neurons. ~\citep{zhang2024tc} introduces a Two-Compartment Leaky Integrate-and-Fire spiking neuron model (TC-LIF) that effectively learns long-term temporal dependencies, demonstrating superior performance in temporal classification tasks and high energy efficiency. Zhang et al ~\citep{zhang2022minicolumn} used SNN to implement a episodic memory model by modeling hippocampal CA1 neurons and microcolum structures. ~\citep{li2023learning} proposes a multi-compartment spiking neural network that enhances information interaction between deep and shallow layers using a Multi-Compartment LIF neuron and Adaptive Gating Unit, achieving improved performance on neuromorphic datasets CIFAR10-DVS and N-Caltech101.

There have been some explorations into using SNNs to implement reinforcement learning algorithms. The PopSAN~\citep{tang2020deep,tang2020reinforcement} built a population-coded spiking actor network and is hybrid trained with a deep critic network to apply SNNs in continuous control tasks. In addition, SNNs are inherently challenging to be optimized, and the dynamical learning environment makes it difficult to train the SNNs directly in DRL environment. To avoid this, Tan et.al~\citep{tan2021strategy} used ANN-SNN conversion method~\citep{li2022efficient,wang2024universal} to convert DQN model to spiking neural networks and achieved decent performance on multiple Atari games. With the surrogate function approximating the spike gradient, the gradient-based optimization method, Spatio-temporal Backpropagation (STBP), is used to train the Spikng-DQN model directly~\citep{liu2021human,chen2022deep}. ~\citep{zhang2021distilling} combined STBP with  knowledge distillation to train SNN in RL task, and achieved significantly accelerating convergence in training process. Sun et.al~\citep{sun2022solving} proposed PL-SDQN model introducing a potential-based layer normalization (pbLN) method to directly train spiking deep Q networks, which solved the spiking vanish problem in the deep spiking convolutional layer of SNNs in DRL. The PL-SDQN achieved superior performance on Atari games compared to existing ANN-SNN conversion methods and other SDQN approaches.
\begin{figure}[!htb]
	\centering
	\includegraphics[width=0.9\linewidth]{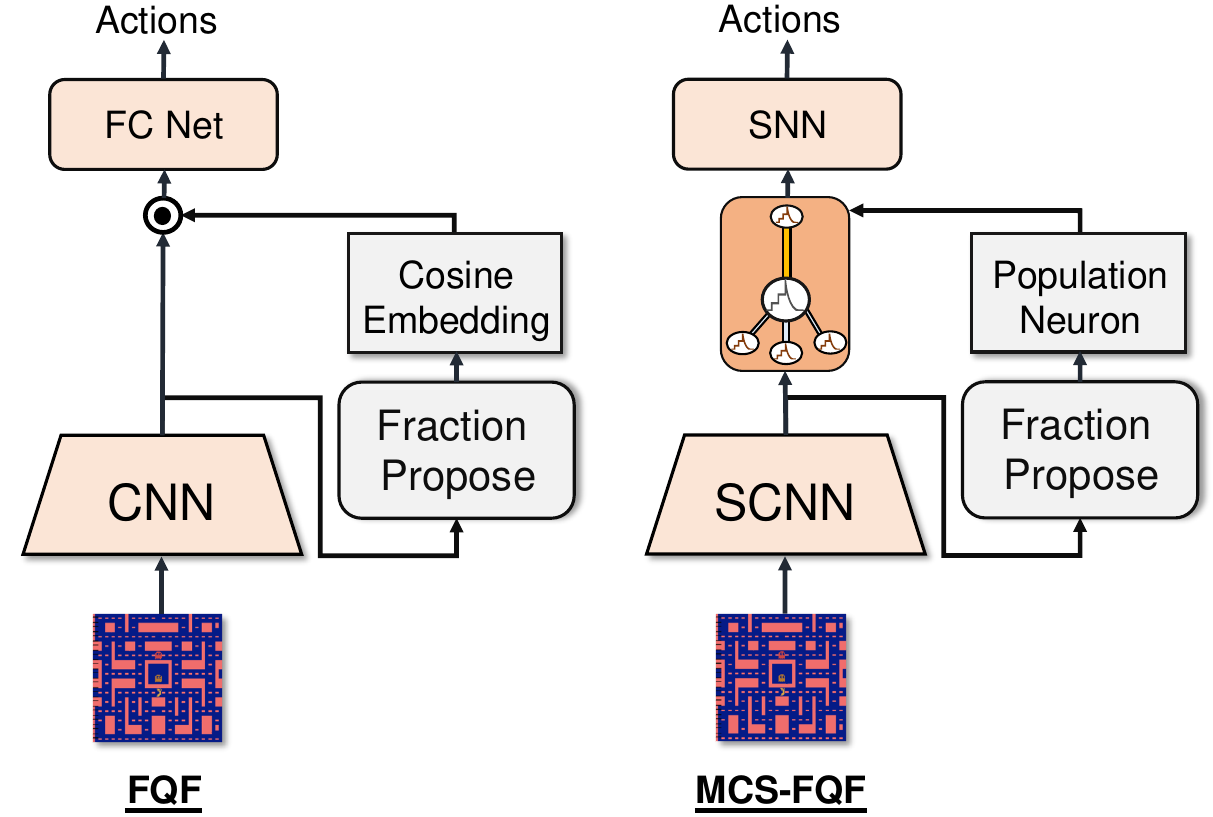} 
	\caption{Network architectures for FQF and MCS-FQF.}
	\label{Fig:fqfcompare}
\end{figure}

\section{Methods}
Implementing distributional reinforcement learning algorithms using spiking neural networks, particularly deep distributional reinforcement learning algorithms based on quantile regression methods, presents several challenges. The first challenge lies in accurately representing continuous quantile fraction values, ranging from 0 to 1, within the spiking neural framework. The second challenge involves the fusion of state information and quantile information, both of which are represented as spike sequences. To address these challenges, this study proposes a multi-compartment neuron model that integrates information through a dendritic structure. In this model, the basal dendrites and apical dendrites of neurons receive spike embeddings corresponding to the observation state and quantile fraction, respectively. These dendritic potentials are then transmitted to the soma, where feature integration occurs. Furthermore, the spiking activity of a population of neurons is utilized to accurately represent continuous quantile values.
Building on this foundation, we implemented the fully spiking FQF model, referred to as the MCS-FQF model. A comparison of network architectures is illustrated in Figure \ref{Fig:fqfcompare}. The MCS-FQF model incorporates a spiking convolutional neural network to extract visual features, while population neurons encode quantile fractions into spike sequences. In contrast to the multiplication method used in the FQF model, the MCS-FQF model leverages the multi-compartment neuron structure to integrate both observation state features and quantile coding features.

\subsection{Multi-compartment Neuron Model for Information Integration}

Unlike conventional artificial neuron models such as Sigmoid and ReLU, spiking neurons in SNNs process input signals by dynamically modulating their membrane potential. When this potential exceeds a certain threshold, the neuron emits a spike as its output. Although the dynamic behavior of spiking neurons enhances the computational capacity of the network, it often fails to fully exploit the structural properties of biological neurons. Moreover, the specific contributions of various neuronal components to the network's computational power, particularly in complex tasks, remain largely unexplored and merit further investigation.

The LIF neuron is a point structures computing model described by Eq.~(\ref{EQ:LIF}). It only uses the somatic membrane potential $u_t$ to process the input signal $x_t$. When $u_t$ exceeds the threshold $V_{th}$, the neuron fires a spike as $s_t=1$, otherwise $s_t=0$ for no spikes. 
\begin{equation}
	\left\{\begin{array}{l}
		\tau_L\frac{du_t}{dt} = -u_t+ x_t,  \\
		s_t= 1, u_t \gets V_{reset}\quad \text{if}\ u_t>V_{th}.
	\end{array}\right.
	\label{EQ:LIF}
\end{equation}

\begin{figure}[!htb]
	\centering
	\includegraphics[width=1.0\linewidth]{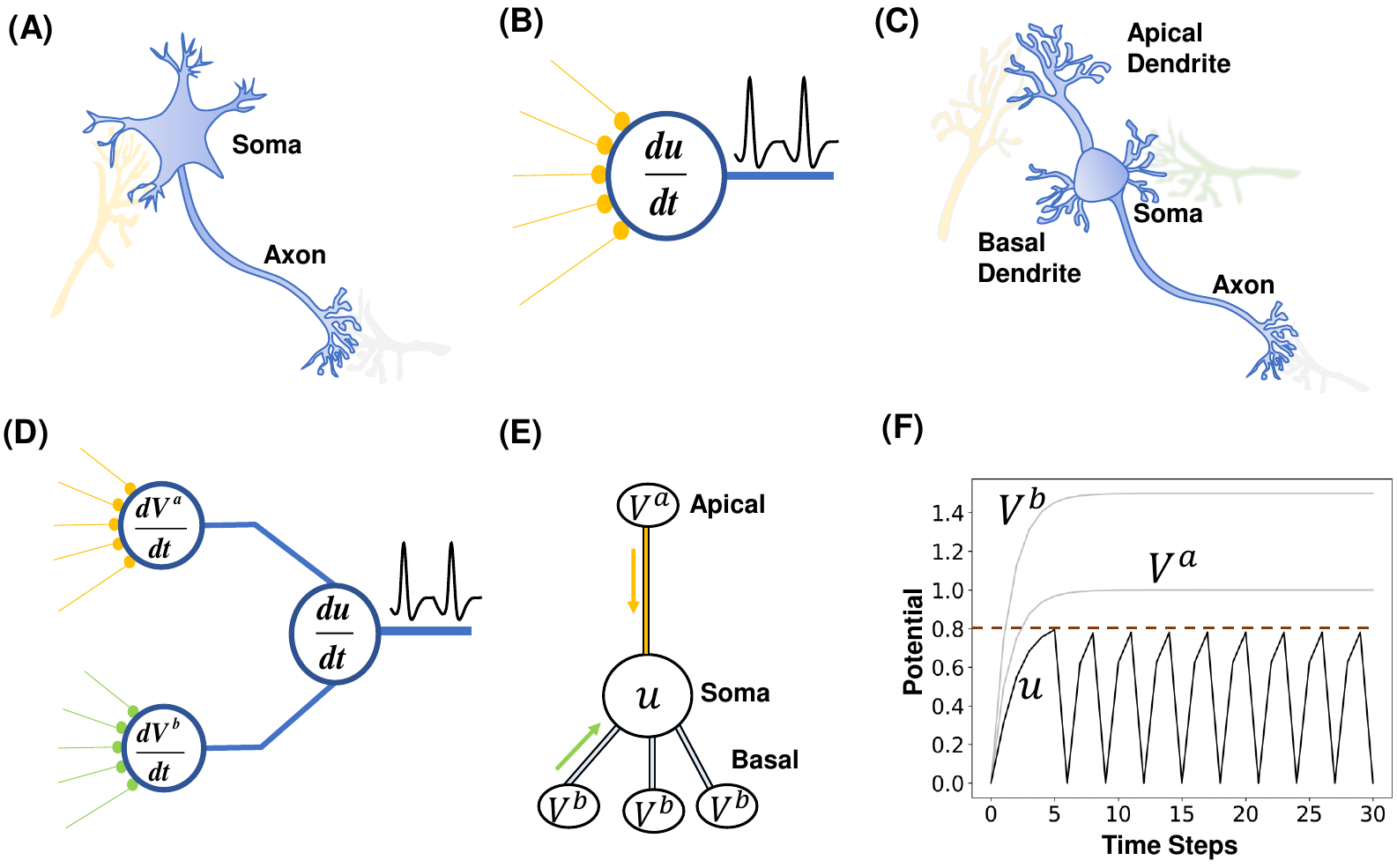} 
	\caption{Point neuron model and multi-compartment neuron model. \textbf{(A)} Point neuron with soma compartment. \textbf{(B)} Computational model for point neuron with single information processing node. \textbf{(C)} Multi-compartment neuron with basal dendrite, apical dendrite and soma. \textbf{(D)} Computational model for multi-compartment neuron. \textbf{(E)} Structure and potentials influence of MCN. \textbf{(F)} Firing activity of MCN model. The potentials are recorded from constant inputs signal with $x_a=1.0$ and $x_b=1.5$ within $T=30$ time steps. The parameters of MCN model are set as $\tau_A=\tau_B=2.0$, $\tau_L=4.0$, $g_A=g_B=g_L=1.0$, and the threshold potential is $V_{th}=0.8$.}
	\label{Fig:neurons}
\end{figure}

Inspired by the dendritic integration function of pyramidal neurons in the hippocampal CA3 region~\citep{makara2013variable}, we developed a multi-compartment neuron model to integrate state and fraction information. Unlike the traditional LIF neuron model, the MCN model features three distinct compartments: the basal dendrite, apical dendrite, and soma, as illustrated in Figure~\ref{Fig:neurons}. The basal and apical dendrite compartments contain weighted synapses and receive inputs from different sources. The apical dendrites, located farther from the soma, receive modulated signals from other brain regions, while the basal dendrites, positioned closer to the soma, primarily process feedforward inputs.

\begin{equation}
	\left\{\begin{array}{l}
		\tau_B\frac{dV_t^b}{dt}=-V_t^b + x_t^b \\
		x_t^b=\sum_jw^b_jo^b_{j,t}
	\end{array}\right.
	\label{EQ:basal}
\end{equation}

The computing process of the basal dendrite is described by Eq.~(\ref{EQ:basal}). The $o^b_{j,t}$ is the spike signal neuron $j$ input to the basal dendrite at time step $t$. It is summed by the weighted synapse $w_b$ to the dendritic postsynaptic potential $x^b_t$. The dendritic potential $V^b_t$ integrates $x^b_t$ with time constant $\tau_B$. 

\begin{equation}
	\left\{\begin{array}{l}
		\tau_A\frac{dV^a_t}{dt}=-V^a_t + x^a_t  \\
		x^a_t=\sum_jw^a_jo^a_{j,t}
	\end{array}\right.
	\label{EQ:apical}
\end{equation}

In our proposed MCN model, the apical dendrites share the same computational structure as the basal dendrites, as outlined in Eq.~(\ref{EQ:apical}). Although apical dendrites may exhibit distinct computational properties, the primary focus of this work is on the neuron's ability to integrate diverse types of information. The MCN model is utilized to process embedding spike signals representing both state and quantile fraction. Both basal and apical dendritic synaptic weights are learnable parameters, allowing the MCN to serve as a flexible and adaptable unit for information integration.

The potentials of basal and apical dendrites are conducted to the soma compartment, and the MCN model integrates this information to the somatic potential $u_t$ as in Eq.~(\ref{EQ:mc}).
\begin{equation}
	\tau_L\frac{du_t}{dt}=-u_t + \frac{g_B}{g_L}(V^b_t-u_t)+\frac{g_A}{g_L}(V^a_t-u_t)
	\label{EQ:mc}
\end{equation}
where the $g_B$, $g_A$ and $g_L$ are constant hyperparameters as basal dendrite conductance, apical dendrite conductance, and leaky conductance, respectively, the $\tau_L$ is the somatic leaky time constant. The neuron fires a spike when the somatic potential $u_t$ exceeds the threshold $V_{th}$. We omit the time step subscripts below for brevity.

\begin{theorem}\label{thm}
	Let $Z=\frac{g_B+g_A+g_L}{\tau_L g_L}$, the somatic potential $u(t)$ is the spatial-temporal integration of the apical and basal dendritic potentials as:
	\begin{equation}
		u(t) = \int^{t}_{-\infty}\frac{e^{Z t}}{\tau_L}[\frac{g_B}{g_L}V^b(t)+\frac{g_A}{g_L}V^a(t)]dte^{-Z t}
		\label{EQ:mc_po}	
	\end{equation}
	\label{Theo:mc_po}
\end{theorem}

\begin{pot} 
	The MCN membrane potential equation is: 
	\begin{equation}
		\tau_L\frac{du}{dt}=-u(t) + \frac{g_B}{g_L}(V^b(t)-u(t))+\frac{g_A}{g_L}(V^a(t)-u(t))
	\end{equation}
	which can be rewritten as: 
	\begin{equation}
		\frac{du}{dt}+Zu(t)+C(t)=0
		\label{Eq:linear}
	\end{equation}
	\begin{equation}
		Z=\frac{g_B+g_A+g_L}{\tau_L g_L}
	\end{equation}
	\begin{equation}
		C(t) = -\frac{1}{\tau_L}[\frac{g_B}{g_L}V^b(t)+\frac{g_A}{g_L}V^a(t)]
	\end{equation}
	This is a first-order linear differential equation, and its general solution is:
	\begin{equation}
		u(t) = A(t)e^{-\int Zdt} =  A(t)e^{-Zt}
		\label{Eq:gen}
	\end{equation}
	Substituting the solution for $u(t)$ of Eq.~(\ref{Eq:gen}) into Eq.~(\ref{Eq:linear}), we get:
	\begin{equation}
		A(t) = \int -e^{Zt}C(t)dt
	\end{equation}	
	Thus, $u(t)$ is solved as:
	\begin{equation}
		u(t) = \int^{t}_{-\infty}\frac{e^{Z t}}{\tau_L}[\frac{g_B}{g_L}V^b(t)+\frac{g_A}{g_L}V^a(t)]dte^{-Z t}
	\end{equation}
\end{pot}

In contrast to point neuron models that combine input signals through simple summation, the MCN model integrates apical and basal dendritic potentials across both spatial and temporal dimensions, as outlined in Eq.~(\ref{EQ:mc_po}). The relationship between somatic and dendritic potentials in the MCN model is depicted in Figure~\ref{Fig:neurons}(E)(F).

\subsection{Population Encoding For Quantile Fraction Implicit Representation} 
\begin{figure}[!htb]
	\centering
	\includegraphics[width=1.0\linewidth]{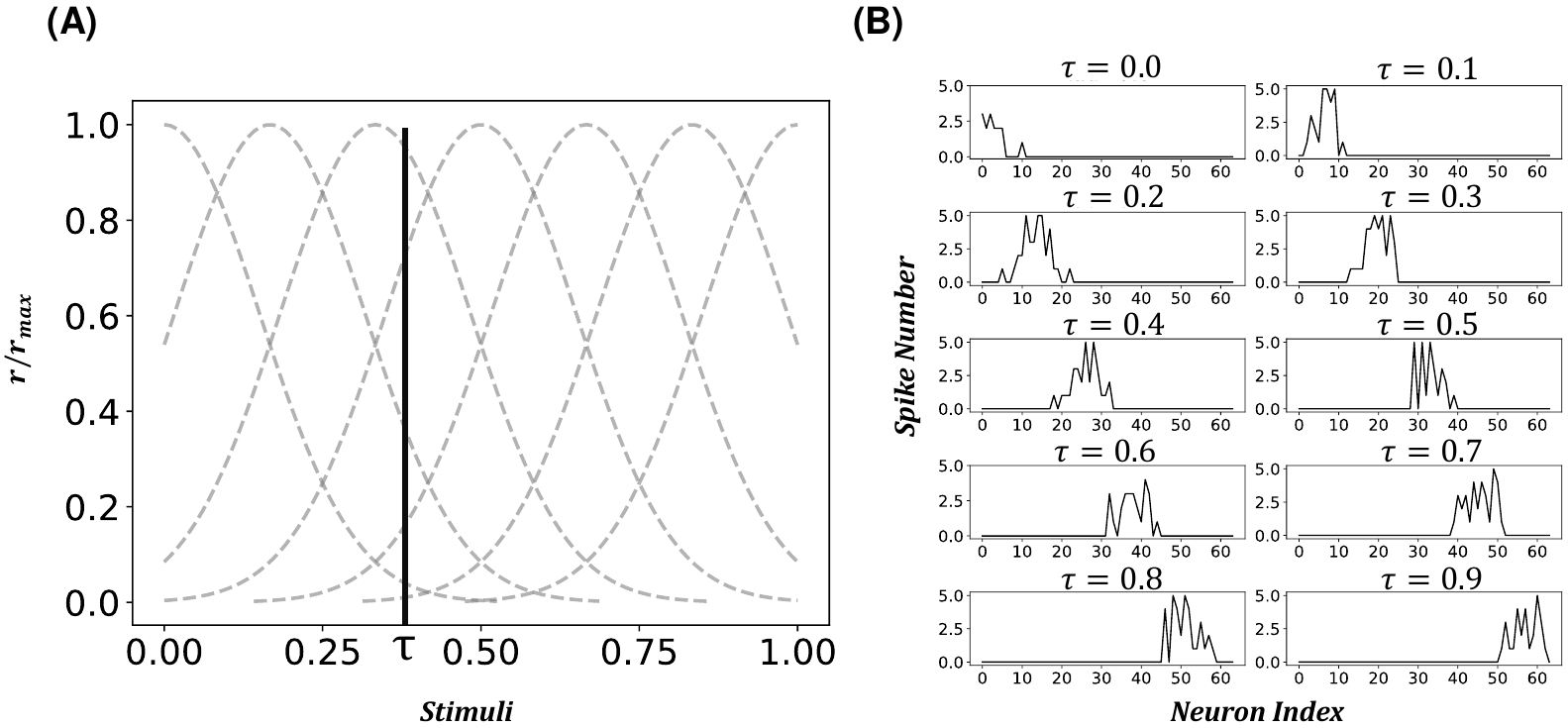} 
	\caption{The spiking activity of population neuron. \textbf{(A)} Population neurons with Gaussian receptive field. The gray dotted line represents the relationship between the firing rate of different neurons and input stimuli. The intersection of the solid vertical line with the activity curve of each neuron is the firing rate of that neuron when encoding the corresponding quantile fraction value. \textbf{(B)} The spike numbers of $M=64$ neurons within  $T=8$ time periods when representing different fraction values.}
	\label{Fig:pop}
\end{figure}

Quantile regression based distributional reinforcement learning model takes the selected quantile fractions $\tau$ as inputs and outputs the corresponding quantile values. The quantile fraction is proposed by a fully parameterized function with the value's range is $\tau\in[0,1]$. To combine SNN with deep distributional reinforcement learning, we propose a population encoding method using the neural spikes as the implicit representation of quantile fraction. We use $M$ neurons with stimulus selective sensitivity as the encoding neuron populations. Each neuron has a Gaussian receptive field, and the relationship between the $j_{th}$ neuron's spike firing rate $r_j$ and its input stimulus $s_j$ is:

\begin{equation}
	r_j = \phi_j exp(-\frac{(s_j-\mu_j)^2}{2\sigma^2_j})
	\label{EQ:gussian}
\end{equation}
where $\mu_j$ is the input stimulus which is most consistent with the neuron's expectations and is also the largest firing rate point. And the $\sigma_j$ determines the range of the receptive field. The smaller the value of $\sigma_j$, the more selective the neuron is for the input stimulus, as shown in Figure~\ref{Fig:pop}(A).

To represent the quantile fractions with population neural spikes, we encode the $i_{th}$ fraction value $\tau_i$ to the collaborative activities of the $M$ neurons. The firing rate of the $j_{th}$ neuron for the value $\tau_i$ is: 
\begin{equation}
	r_{ij} = \phi_j exp(-\frac{(\tau_i-\mu_j)^2}{2\sigma^2_j})
	\label{EQ:en_ij}
\end{equation}
and each neuron has a unique preferred stimulus with $\mu_i=\frac{i}{N}$. Because the quantile fractions are homogeneous values ranging from 0 to 1, all neurons have the same receptive characteristic as:

\begin{equation}
	\forall i\in[1, 2,.., M], \sigma_i=C
	\label{EQ:pop_sigma}
\end{equation}

The neurons' spiking activity is a Poisson process.  And the quantile fraction $\tau_i$ is encoded to population spike $S_{\tau_i}$ as Eq.~(\ref{EQ:pop_tau}). The neuron $j$ fires $n$ spikes within time interval $\Delta t$ with possibility $P_{ij,\Delta t}[n]$.
\begin{equation}
	S_{\tau_{i,j}}\sim P_{ij,\Delta t}[n] = \frac{(r_{ij}\Delta t)^2}{n!}exp(-r_{ij}\Delta t)
	\label{EQ:pop_poisson}
\end{equation}

\begin{equation}
	S_{\tau_{i}} = \{S_{\tau_{i,j}}, j=\{1, 2, ..., M\}\}
	\label{EQ:pop_tau}
\end{equation}

\begin{figure}[!htb]
	\centering
	\includegraphics[width=1.0\linewidth]{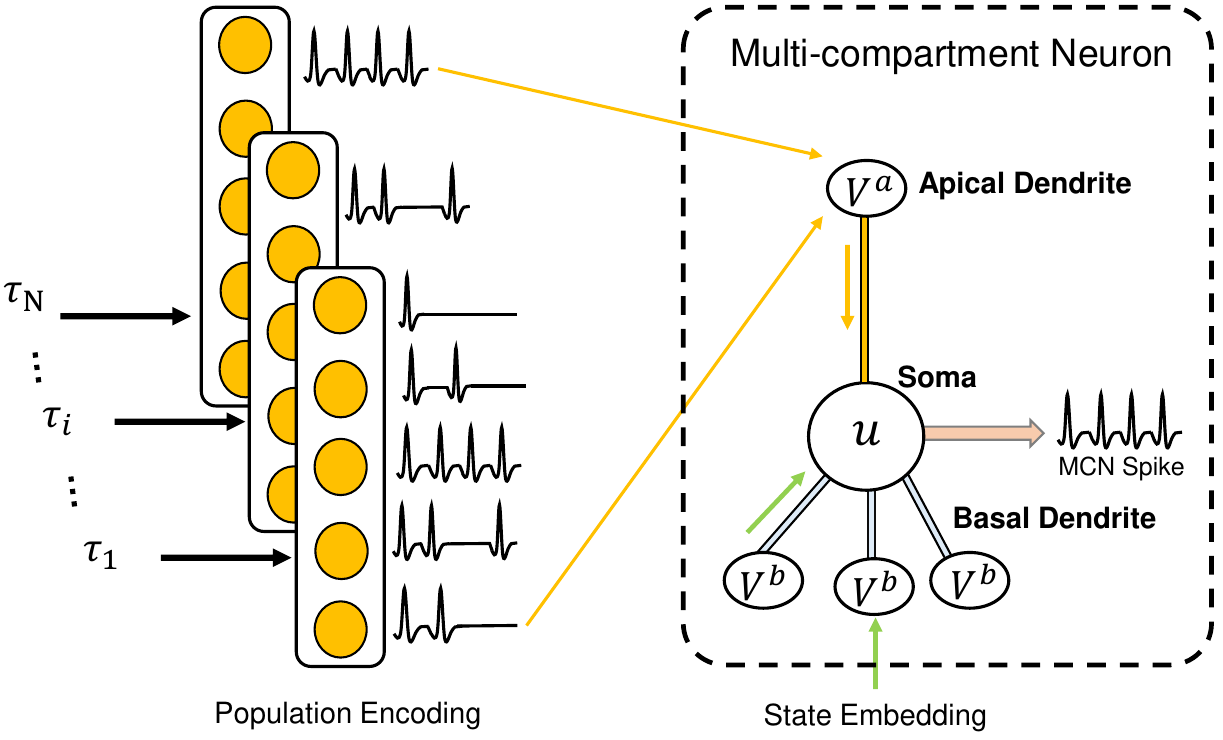} 
	\caption{Encoding quantile fraction value with population neuron.}
	\label{Fig:pop_neuron}
\end{figure}

We utilize $M=64$ neurons to encode fraction values. Figure~\ref{Fig:pop}(B) illustrates the firing activity of the population neurons during the encoding of different fraction values. The spiking implicit representation of the fraction value is input to the apical dendrite, where the apical dendritic potential  $V^a$ is integrated with the observation state information in the soma of the multi-compartment neuron, as depicted in Figure~\ref{Fig:pop_neuron}.

\subsection{Multi-compartment Spiking-FQF Model}

\begin{figure}[!htb]
	\centering
	\includegraphics[width=1.0\linewidth]{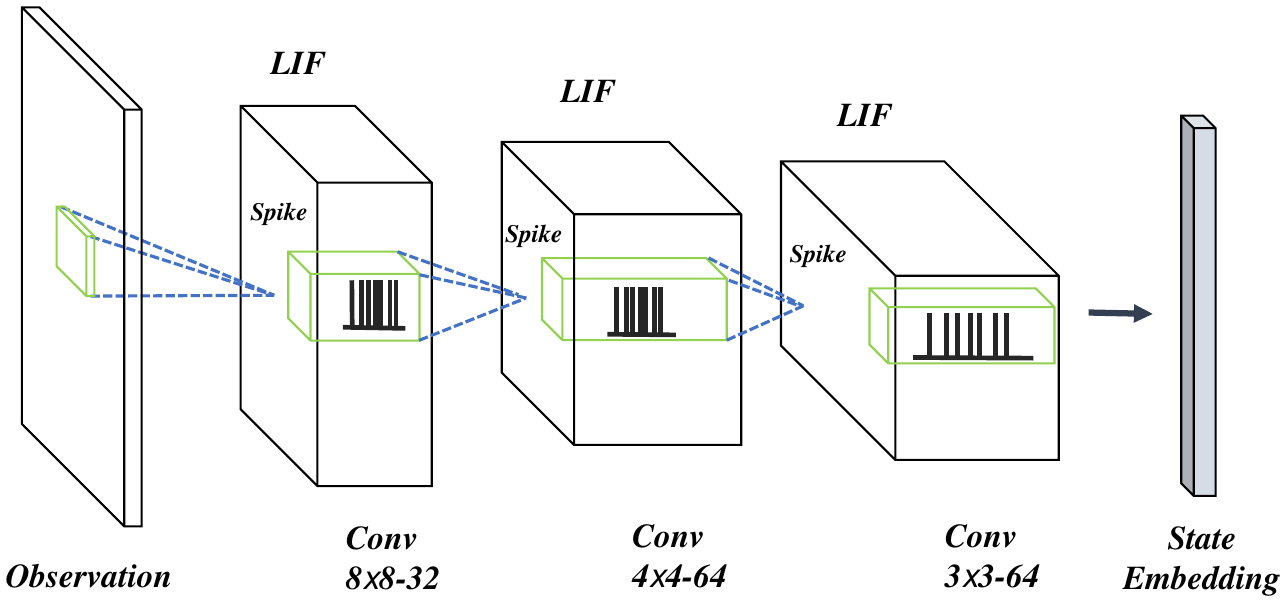} 
	\caption{Spiking convolutional neural network used for encoding image observation to spike state embedding. The note "8x8-32" means that the convolution layer has the structure of "8x8" kernel size and 32 feature channels.}
	\label{Fig:scnn}
\end{figure}

\begin{figure*}[!htb]
	\centering
	\includegraphics[width=0.95\linewidth]{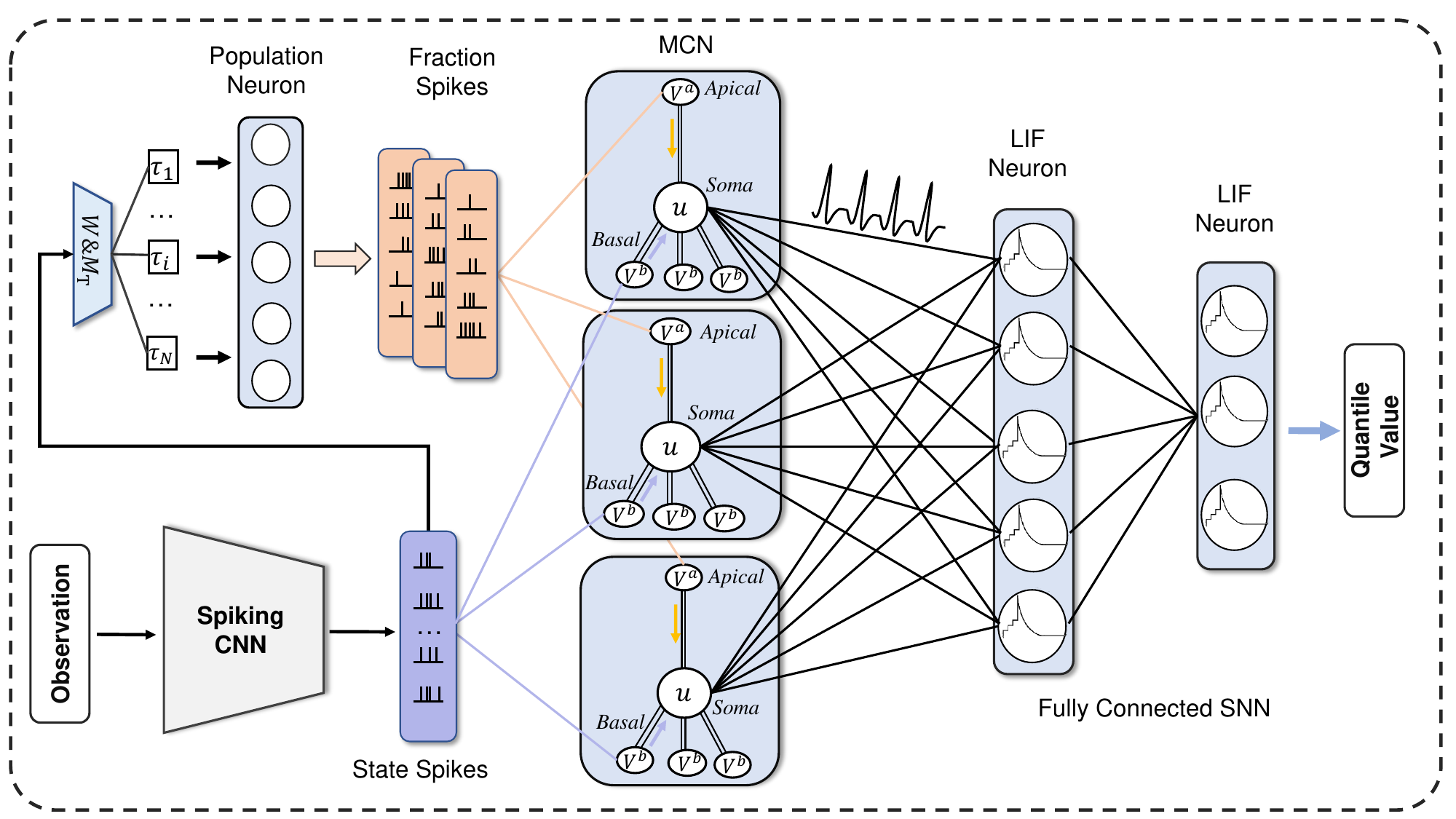} 
	\caption{The framework of MCS-FQF. Multi-compartment neurons receive the state embedding inputs from spiking convolution neural network and the quantile fraction encoding inputs from population neurons. The somatic spikes of MCN input to a fully connected SNN to generate the quantile value.}
	\label{Fig:mcs-fqf}
\end{figure*}

We propose the MCS-FQF model, based on the multi-compartment neuron and population encoding method, as illustrated in Figure~\ref{Fig:mcs-fqf}. To the best of our knowledge, this is the first work to apply spiking neural networks to deep distributional reinforcement learning. The MCS-FQF model utilizes a spiking convolutional neural network to extract state embedding spike trains from image observations, as depicted in Figure~\ref{Fig:scnn}. The SCNN, constructed with LIF neurons, consists of three convolutional layers, producing the spike embedding of the state $O^s$ from its outputs.

The number of quantile fractions is set to  $N$, with  $\tau_0=0$, $\tau_N=1$, and for $i=[1,...,N-1]$, $\tau_{i-1}<\tau_i$, following the approach in~\citep{yang2019fully}. The quantile fractions are calculated using the cumulative distribution probability generated from  $O^s$, as described in Eq.~(\ref{EQ:cal_phi}-\ref{EQ:cal_tau}). The state embedding is then processed by the learnable weight  $W^f$, and the mean activities of SCNN neurons within a time window $T$ are treated as logits for the quantile fraction probability.

\begin{equation}
	\phi_i = \frac{1}{T}\sum_{t=0}^{T-1} W^f_iO^s_{t}
	\label{EQ:cal_phi}
\end{equation}

\begin{equation}
	p_k=\frac{exp(\phi_k)}{\sum_j exp(\phi_j)}
	\label{EQ:p_k}
\end{equation}

\begin{equation}
	\tau_i = \sum_{k=0}^{i-1}p_k
	\label{EQ:cal_tau}
\end{equation}

Multi-compartment neurons integrate state and quantile fraction embeddings as described in Eq. (\ref{EQ:basal_state}-\ref{EQ:mc_fuse}). Apical dendrites receive spike representations of the quantile fractions, while basal dendrites process the state embedding inputs. The resulting somatic potential $u$ is derived from the integration of the basal dendritic potential $V_s^b$ and the apical dendritic potential $V_\tau^a$.

\begin{equation}
	\left\{\begin{array}{l}
		\tau_B\frac{dV_s^b}{dt}=-V_s^b + x^b \\
		x^b=\sum_jw^b_jO^s_{j}
	\end{array}\right.
	\label{EQ:basal_state}
\end{equation}

\begin{equation}
	\left\{\begin{array}{l}
		\tau_A\frac{dV_\tau^a}{dt}=-V_\tau^a + x^a  \\
		x^a=\sum_jw^a_jS_{\tau,j}
	\end{array}\right.
	\label{EQ:apical_tau}
\end{equation}

\begin{equation}
	\tau_L\frac{du}{dt}=-u + \frac{g_B}{g_L}(V_s^b-u)+\frac{g_A}{g_L}(V_\tau^a-u)
	\label{EQ:mc_fuse}
\end{equation}

When the somatic potential  $u_t$ surpasses the threshold potential $V_{th}$, the MCN emits spike trains denoted as  $S^{m}(t)$. Subsequently, these spikes are processed by a two-layer fully connected spiking neural network  composed of  LIF neurons. The quantile values $F^{-1}_w(\tau_i)$ are estimated from the output spikes of this final SNN, as detailed in Eq.~(\ref{EQ:quantile}). The weight parameter $W^L$ is optimized alongside the entire network. The comprehensive computational procedure of the MCS-FQF model is outlined in Algorithm~\ref{alg:Proc}, which generates the state-action values as described in Eq.~(\ref{EQ:qvalue}).

\begin{algorithm}[tb]
	\caption{Proceeding process of MCS-FQF.}
	\label{alg:Proc}
	\textbf{Input}: Observation state $S_t$  \\
	\textbf{Parameter}: Decay factor $\tau_L$, $\tau_A$, $\tau_B$; conductance $g_A$, $g_B$, $g_L$; fraction number $N$, population  number $M$; simulation time-window $T$; membrane potential threshold $V_{th}$; layers number $L$. \\
	\textbf{Output}: State-action values $Q$. \\
	\begin{algorithmic}[1] 
		\STATE Initialize neuron weight $w^l$ and potential $u^l_0$  $\forall \ l=0,1,2,..., L-1$. \\
		\STATE \textcolor[rgb]{0.5, 0.5, 0.5}{// Compute the state embedding}
		\FOR{$t=0$ to $T-1$}	
		\STATE	$O^s_t = SCNN(S_t)$ 
		\ENDFOR
		\STATE 	\textcolor[rgb]{0.5, 0.5, 0.5}{// Compute the fractions}
		\FOR{$i=1$ to $N$}
		\FOR{$t=0$ to $T-1$}	
		\STATE	$\tau_i=F(W^fO^s_t)$ as in Eq.~(\ref{EQ:cal_phi}-\ref{EQ:cal_tau})
		\ENDFOR
		\ENDFOR
		\STATE 	\textcolor[rgb]{0.5, 0.5, 0.5}{// Compute the quantile values $F^{-1}_w$}
		\FOR{$t=0$ to $T-1$}	
		\STATE \textcolor[rgb]{0.5, 0.5, 0.5}{ // Calculate fraction embedding by population}
		\STATE \textcolor[rgb]{0.5, 0.5, 0.5}{ // encoding}
		\FOR{$j=0$ to $M-1$}
		\STATE	 $S_{\tau_j,t}=LIF(PopEmb(\tau_j))$ 
		\ENDFOR
		\FOR{$i=0$ to $N-1$}
		\STATE	$F^{-1}_w(\tau_i, t)=SNN(MCN(S_{\tau_j,t}, O^s_t))$ 
		\ENDFOR
		\ENDFOR
		\STATE Calculate state-action value $Q$ as Eq.~(\ref{EQ:qvalue}).
		\STATE \textbf{return} $Q$
	\end{algorithmic}
\end{algorithm}

\begin{equation}
	F^{-1}_w(\tau_i) = \frac{1}{T}\sum_{t=0}^{T-1} W^LO^L_{i,t}
	\label{EQ:quantile}
\end{equation}

\begin{equation}
	Q(s,a) = \sum_{i=0}^{N-1}(\tau_{i+1}-\tau_i)F^{-1}_w(\hat{\tau}_i)
	\label{EQ:qvalue}
\end{equation}

\begin{equation}
	\hat{\tau}_i=\frac{\tau_{i}+\tau_{i+1}}{2}
	\label{EQ:tau_hat}
\end{equation}

The MCS-FQF model is trained using the Spatio-Temporal Backpropagation  algorithm~\citep{Wu2018Spatio}. The weights of both the spiking convolutional neural network and the final fully connected spiking neural network are optimized by minimizing the Huber quantile regression loss~\citep{huber1992robust}.

\begin{equation}
	\mathbb{HL}(\tau,\delta_{ij}) = |\tau-(1-\mathbb{H}(\delta_{ij}))|\frac{\mathbb{L}(\epsilon,\delta_{i,j})}{\epsilon}
	\label{EQ:qr}
\end{equation}

\begin{equation}
	\mathbb{L}(\epsilon,\delta_{i,j})=\left\{\begin{array}{ll}
		\frac{1}{2}\delta^2_{ij}, & \text{if}~|\delta_{ij}|\le\epsilon \\
		\epsilon(|\delta_{ij}|-\frac{1}{2}\epsilon) & \text{otherwise}.
	\end{array}\right.
	\label{EQ:hubber}
\end{equation}

\begin{equation}
	\mathbb{H}(\delta_{ij})=\left\{\begin{array}{ll}
		0, &\text{if}~\delta_{ij}\le 0 \\
		1, &\text{otherwise}.
	\end{array}\right.
	\label{EQ:heside}
\end{equation}

where $\delta_{ij}$ is the temporal difference (TD) error for quantile value distribution:

\begin{equation}
	\delta^t_{ij}=r_t+F^{(-1)}(\hat{\tau}_i|s_{t+1},a_{t+1})-F^{(-1)}(\hat{\tau}_j|s_{t},a_{t})
\end{equation}

When neurons fire spikes, the LIF and MCN models are non-differentiable. We use the surrogate function to approximate the gradient of the neuron's spike $o_t$ as:
\begin{equation}
	\frac{\partial o_{t}}{\partial u_{t}} = \frac{2\tau_L}{4+(\pi \tau_L u_{t})^2}
	\label{EQ:surrogate}
\end{equation}

The fraction proposing weight $W^f$ is optimized by minimizing the Wasserstein loss of quantile value distribution:
\begin{equation}
	\mathbb{WL}(\tau)=\sum_{i=0}^{N-1}\int_{\tau_i}^{\tau_{i+1}}|F^{-1}(\theta)-F^{-1}(\hat{\tau}_i)|d\theta
	\label{EQ:wl}
\end{equation}

\begin{equation}
	\frac{\partial \mathbb{WL}(\tau)}{\partial \tau_i} = [2F^{-1}(\tau_i)-F^{-1}(\hat{\tau}_i)-F^{-1}(\hat{\tau}_{i-1})]
\end{equation}

We get the derivative of $\mathbb{WL}(\tau)$ with the fraction proposing weight:

\begin{align}
	\frac{\partial \mathbb{WL}(\tau)}{\partial W^f_i} &= \sum_{n=1}^{N}\frac{\partial \mathbb{WL}(\tau)}{\partial \tau_n}\frac{\partial \tau_n}{\partial \phi_i}\frac{\partial \phi_i}{W^f_i} \nonumber \\
	& = \frac{1}{T}\sum_{n=1}^{N}\sum_{k=0}^{n-1}\sum_{t=0}^T\frac{\partial \mathbb{WL}(\tau)}{\partial \tau_n}\Delta_{i,k}O^s_t 
\end{align}

\begin{equation}
	\Delta_{i,k}=(-p_kp_i)+(N-i+1)p_i(1-p_i)
\end{equation}

Basal dendrite synaptic weights $w^b$ and apical dendrite synaptic weight $w^a$ are trained with error signal $\delta_{o_t}=\frac{\partial \mathbb{HL}(\tau,\delta_{ij}) }{\partial o_t}$ back propagating from the fully connected SNN. We get the derivative of the MCN basal dendrite synaptic weight and apical dendrite synaptic weight as Eq.~(\ref{EQ:basal_der}) and Eq.~(\ref{EQ:apical_der}) respectively.

\begin{small}
\begin{align}
	&\frac{\partial \mathbb{HL}(\tau,\delta_{ij}) }{\partial w^b} = \frac{\partial \mathbb{HL}(\tau,\delta_{ij}) }{\partial o_t} \frac{\partial o_{t}}{\partial u_{t}} \frac{\partial u_{t}}{\partial V^b_{s,t}} \frac{\partial V^b_{s,t}}{\partial w^b} \nonumber \\
	&=\sum_{t=1}^{T}\sum_{k=0}^{t-1}\delta_{o_t}\frac{2\tau_L}{4+(\pi \tau_L u_{t})^2}m^{T-t}\frac{g_B}{g_L\tau_L}(1-\frac{1}{\tau_B})^{t-k-1}\frac{1}{\tau_B}O_k^s  \nonumber \\
	&=\sum_{t=1}^{T}\sum_{k=0}^{t-1}\frac{2g_B\delta_{o_t} m^{T-t}}{g_L\tau_B(4+(\pi \tau u_{t})^2)}(1-\frac{1}{\tau_B})^{t-k-1}O_k^s
	\label{EQ:basal_der}
\end{align}
\end{small}

\begin{small}
\begin{align}
	&\frac{\partial \mathbb{HL}(\tau,\delta_{ij}) }{\partial w^a} = \frac{\partial \mathbb{HL}(\tau,\delta_{ij}) }{\partial o_t} \frac{\partial o_{t}}{\partial u_{t}} \frac{\partial u_{t}}{\partial V^a_{\tau,t}} \frac{\partial V^a_{\tau,t}}{\partial w^a} \nonumber \\
	&= \sum_{t=1}^{T}\sum_{k=0}^{t-1}\delta_{o_t}\frac{2\tau_L}{4+(\pi \tau_L u_{t})^2}m^{T-t}\frac{g_A}{g_L\tau_L}(1-\frac{1}{\tau_A})^{t-k-1}\frac{1}{\tau_A}S_{\tau, k}  \nonumber \\
	&= \sum_{t=1}^{T}\sum_{k=0}^{t-1}\frac{2g_A\delta_{o_t} m^{T-t}}{g_L\tau_A(4+(\pi \tau_L u_{t})^2)}(1-\frac{1}{\tau_A})^{t-k-1}S_{\tau, k}
	\label{EQ:apical_der}
\end{align}
\end{small}

\begin{equation}
	m = 1 - \frac{1}{\tau_L} - \frac{g_B}{g_L\tau_L} - \frac{g_A}{g_L\tau_L}
\end{equation}

\section{Experiments}
We compared the proposed MCS-FQF model against the vanilla ANN-based FQF model and the ANN-to-SNN conversion method across various Atari games. Experimental results demonstrate that our SNN-based MCS-FQF model outperforms both the FQF baseline and the ANN-SNN conversion approach, underscoring the performance benefits of directly training multi-compartment spiking neural networks. Furthermore, comparisons with state-of-the-art (SOTA) SNN-based deep reinforcement learning methods show that the MCS-FQF model achieves superior performance across the Atari games. An ablation study further confirms that the inclusion of the MCN model and population encoding not only enhances performance but also reduces energy consumption.

\subsection{Performance of directly training multi-compartment spiking neural network}
\begin{figure*}[!htb]
	\centering
	\includegraphics[width=1.0\linewidth]{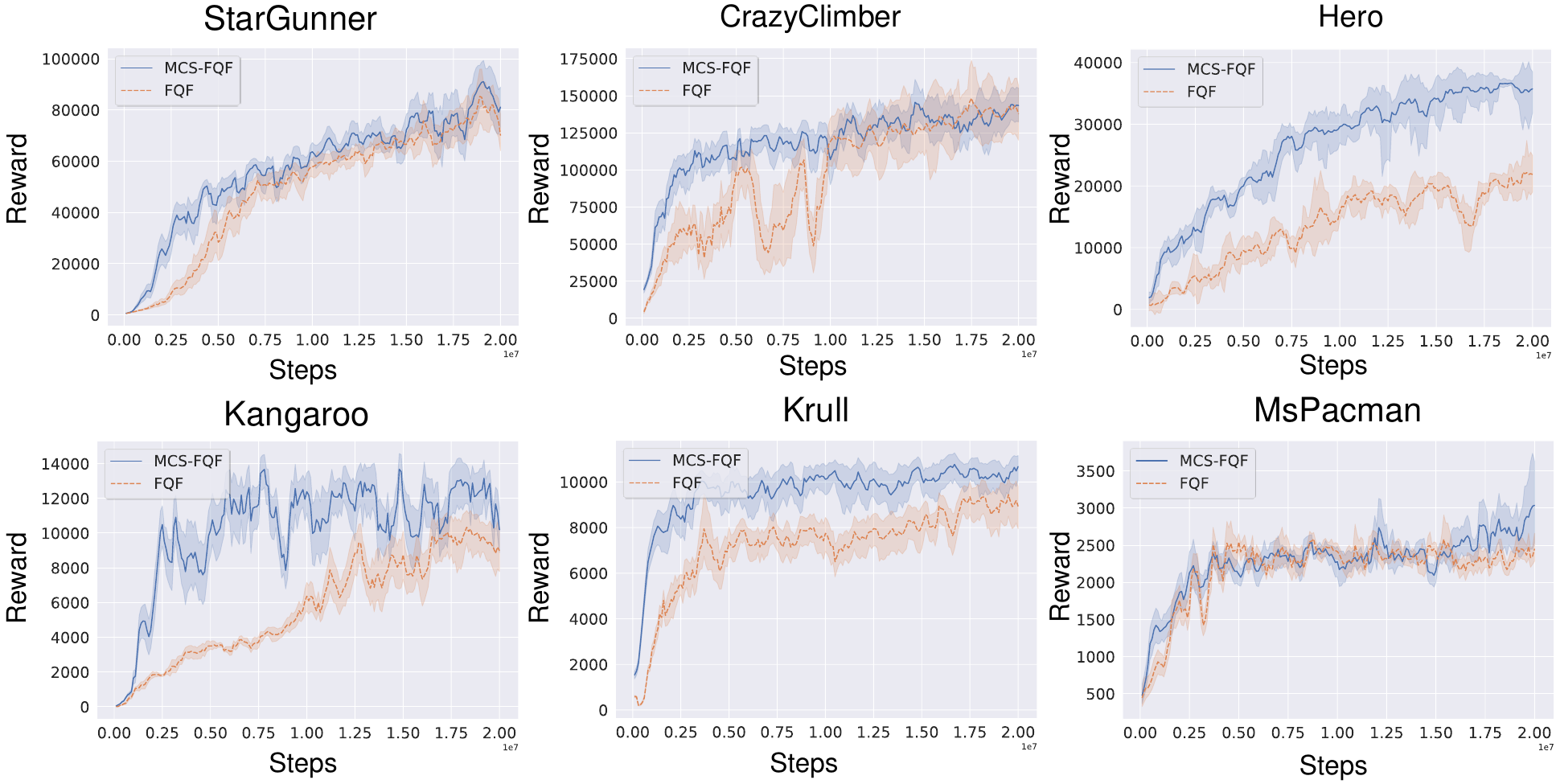} 
	\caption{The performance curve on Atari games of MCS-FQF and baseline FQF models.}
	\label{Fig:result}
\end{figure*}

To demonstrate the superiority of our proposed model, we conducted experiments on Atari games, comparing it with the vanilla FQF model~\citep{yang2019fully} and the Spiking-FQF model implemented using the ANN-SNN conversion method~\citep{li2022efficient}. Unlike the FQF model, which is based on DNNs, the MCS-FQF model incorporates more biologically plausible mechanisms, such as multi-compartment neurons and population encoding, and is trained directly using the surrogate gradient method. These comparative experiments effectively highlight the advantages of our brain-inspired model in enhancing information representation and integration.

The MCS-FQF model employs 512 multi-compartment neurons to integrate state and quantile fraction information. The final fully connected network includes a hidden layer of 512 neurons and outputs $N=32$ corresponding quantile values for each state-action pair. The MCS-FQF model is simulated over  $T=8$ time steps and is trained directly using the STBP algorithm with a gradient surrogate function.

All models were trained under the same experimental conditions, detailed in Table~\ref{Tab:parameter}, for 20 million frame steps. We used the Adam optimizer to minimize the Huber quantile regression loss and the RMSprop optimizer to minimize the Wasserstein loss. Each model was tested in ten independent runs, and the average reward and variance curves for our model and the baseline are presented in Figure~\ref{Fig:result}. The experimental results consistently demonstrate a performance boost with our MCS-FQF model across various games. Notably, in the Hero, Kangaroo, and Krull games, our MCS-FQF model significantly outperforms the baseline FQF model.

\begin{figure*}[!htb]
	\centering
	\includegraphics[width=1.0\linewidth]{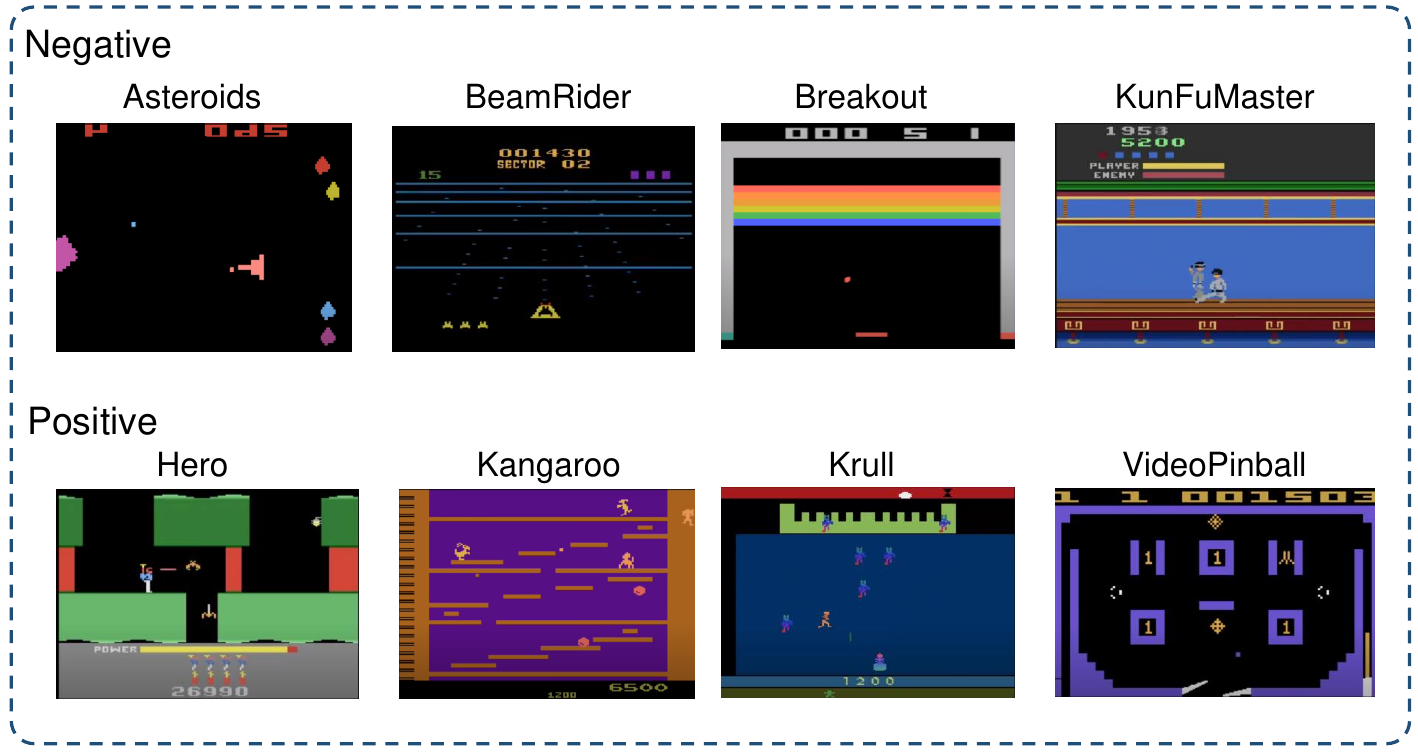} 
	\caption{Game examples used in experiment. The top row shows the game in which the MCS-FQF model scores less than the FQF model. The bottom row shows the games in which the MCS-FQF model has achieved significant performance advantages.}
	\label{Fig:game_exa}
\end{figure*}

The likely reason for this is that these task scenarios are highly complex, involving enemies with intricate shapes and behavior patterns, requiring the agent to make strategic decisions about whether to attack or escape. Additionally, the game environments change rapidly over time, as illustrated in Figure~\ref{Fig:game_exa}. Such scenarios demand a robust ability to integrate information, a strength exhibited by the proposed MCS-FQF model with its multi-compartment neuron structure, leading to significant advantages over the FQF model. On the other three relatively simple tasks, which feature fixed scenes, uniform enemy forms, and a single behavior option for the agent, our model shows slight performance improvements and accelerates the learning process.

\begin{table}[]
	\centering
	\resizebox{.95\columnwidth}{!}{
		\begin{tabular}{l l l l l l}
			\toprule[2pt]
			Parameter & Value & Description & Parameter & Value & Description \\
			\midrule[1pt]
			$\tau_L$  & 2.0  & Decay constant &$g_L$ & 1.0  nS & Leaky conductance  \\
			$V_{th}$  & 1.0 mV & Threshold potential &$g_B$ & 1.0  nS & Basal conductance \\
			$V_{reset}$ & 0.0 mV & Reset potential  &$M$ &64 &Population number  \\
			$T$  & 8 ms & Simulation Times &$C$ &0.05 &receptive constant \\
			$\tau_A$ & 2.0  & Apical constant  &$lr_a$ & $1.0e^{-4}$ & Adam learning ratee\\
			$\tau_B$ & 2.0  & Basal constant &$lr_f$& $2.5e^{-9}$&RMSprop learning rate\\
			$g_A$ & 1.0 nS  & Apical conductance  &$N$ &32 &Fraction number \\
			\bottomrule[2pt]
	\end{tabular}}
	\caption{Parameter settings for experiments.}
	\label{Tab:parameter}
\end{table}

Since no prior work has applied spiking neural networks or multi-compartment neuron models to deep distributional reinforcement learning tasks, we opted to use the ANN-SNN conversion method to transform the ANN in the FQF model into an SNN, thereby implementing the Spiking-FQF model. This conversion method is among the most stable and generalized approaches for applying deep SNNs to new tasks. We converted the trained FQF model into a Spiking-FQF model using the state-of-the-art ANN-SNN conversion technique proposed in~\citep{li2022efficient}, with a simulation time of $T_c=256$ ms, ensuring the high performance of the Spiking-FQF model.

\textbf{
	\begin{table*}[!]
		\centering
		\resizebox{1.0\textwidth}{!}{
			\begin{tabular}{l l l l  l l l l l l}	
				\toprule[2pt]
				& \multicolumn{3}{c}{\textbf{FQF}} & \multicolumn{3}{c}{\textbf{ANN-SNN Conversion}} &\multicolumn{2}{c}{\textbf{MCS-FQF}} \\
				\textbf{Game} & Score & $\pm$std & (Score$\%$)  & Score & $\pm$std &(Score$\%$)   & Score & $\pm$std & (Score$\%$)   \\
				\midrule[1pt]
				Asteroids             & \textbf{2292.0} &812.9  & (35.47$\%$)  &1244.9   &396.4     &(31.84$\%$)    & 1666.0 & 718.4  &(43.12$\%$)      \\
				Atlantis              & 2854490.0       &40015.4    & (1.40$\%$)  &1571086.9   &29768.5     &(1.89$\%$) & \textbf{3267690.0} & 120276.6  &(3.68$\%$) \\
				BeamRider    		  &\textbf{18192.0} &9300.4   & (51.12$\%$)  &8227.6   &3147.5     &(38.25$\%$)  & 15302.2   & 6990.5   &(45.68$\%$) \\
				Bowling               & 59.7            &0.9     & (1.50$\%$)  &37.7   &3.1     &(8.29$\%$) & \textbf{63.5}   & 8.5   &(13.46$\%$) \\
				Breakout     		  &\textbf{592.0}           &203.5      & (34.38$\%$)   &295.4   &90.1     &(30.51$\%$)  &501.5      & 170.3      &(33.96$\%$) \\
				Centipede    		  & 7106.7    		&3617.9    & (50.91$\%$)	 &5015.8   &2199.2     &(43.84$\%$)  & \textbf{7375.9}    & 2929.6  &(39.72$\%$) \\
				CrazyClimber 		  &157910.0         &43971.0   & (27.84$\%$)	& 2440.0  & 874.3    &(35.83$\%$) & \textbf{163010.0}    & 21402.8  &(13.13$\%$) \\
				Enduro                &3421.2           & 1274.0  & (37.24$\%$)	 &2019.0   &591.3     &(29.29$\%$)  & \textbf{4156.4}    & 874.7  &(21.04$\%$) \\
				Frostbite             &7594.0           & 825.3     & (10.87$\%$)     &5269.9   &797.9     &(15.14$\%$)       & \textbf{8733.0}     &933.6  &(10.69$\%$) \\
				Hero                  & 25709.5         & 16881.6  & (65.66$\%$)	 &16851.7   &1389.7     &(8.25$\%$)  & \textbf{36912.5}    & 1091.4  &(2.96$\%$) \\
				Kangaroo	  		  & 11520.0         & 1647.3    & (14.30$\%$)   &7488.6   &1248.4     &(16.67$\%$)   & \textbf{15080.0}     & 325.0    &(2.15$\%$) \\
				Krull	      		  & 10207.0         & 773.0   & (7.57$\%$) &6490.2   &820.2     &(12.64$\%$)   & \textbf{11276.0}     & 599.5    &(5.32$\%$) \\
				KungFuMaster 		  & \textbf{44460.0}     	&11357.1    & (25.54$\%$)  &18725.8   &3506.5     &(18.73$\%$)   & 34820.0     & 9692.9    &(27.84$\%$) \\
				MsPacman	  		  &3398.0 	        & 416.5    & (12.26$\%$)	  &1761.9   &650.2     &(36.90$\%$)  & \textbf{3841.0}	   & 1269.7   &(33.06$\%$) \\
				Qbert                 &  16515.0   & 621.5   & (3.76$\%$)   &8783.0   &422.9     &(4.82$\%$)  & \textbf{18042.5}     & 1577.46   &(8.74$\%$)\\
				RoadRunner            & \textbf{53730.0}    & 7760.4   & (14.44$\%$)  &26593.7   &2458.9     &(9.25$\%$)   & 52790.0     & 13469.5   &(25.52$\%$)\\
				StarGunner            & 97180.0	   & 14331.4   & (14.75$\%$)   &56529.9   &15062.0     &(26.64$\%$) & \textbf{103100.0}     & 17266.19  &(16.75$\%$) \\
				Tutankham             & 267.9	   & 10.5   & (3.93$\%$)   &157.9   &15.9     &(10.04$\%$) & \textbf{298.5}     & 37.0  &(12.41$\%$) \\
				VideoPinball & 357333.9  & 282594.5 & (79.08$\%$)  &244185.7   &203054.5     &(83.16$\%$)  & \textbf{606765.7}    & 250377.9  &(41.26$\%$) \\ 
				\bottomrule[2pt]		
		\end{tabular}}
		\caption{Scores for Atari game experiments. The highest score for each game is bolded.}
		\label{Tab:game_res}
	\end{table*}
}

We conducted comparative experiments on 19 Atari games using the MCS-FQF model, the baseline FQF model, and the ANN-SNN conversion-based Spiking-FQF model. For each model, we recorded the mean scores, standard deviations (std), and the ratio of standard deviation to mean score (Score$\%$) across ten trials, as presented in Table~\ref{Tab:game_res}. The results show that the ANN-SNN conversion-based Spiking-FQF model performs worse than the FQF model across all games, whereas our MCS-FQF model achieves higher scores than the FQF model in most cases. This demonstrates the effectiveness of our SNN-based deep distributional reinforcement learning model. Furthermore, the MCS-FQF model significantly outperforms the conversion-based SNN model, underscoring the effectiveness and robustness of our proposed direct training approach with the MCN model and population encoding method.

As shown in Table~\ref{Tab:game_res}, our model's scores are slightly lower than those of the FQF model in five tasks, though they remain significantly higher than those of the ANN-SNN conversion model. We analyzed the reasons why the MCS-FQF model underperforms the FQF model in these particular games. For instance, in Asteroids, BeamRider, and Kung Master—where there is a notable score difference between the MCS-FQF and vanilla FQF models—the game environments are relatively simple, and the agent has limited action choices, similar to the "left" or "right" options in Breakout. In contrast, the other 14 tasks are more complex, as illustrated in Figure~\ref{Fig:game_exa}, and our model achieves the best performance in these scenarios. Notably, in the Video Pinball game, our model's scores are approximately 1.7  $\times$ higher than those of the FQF model and 2.5 $\times$ higher than those of the ANN-SNN conversion-based Spiking-FQF model. Overall, our MCS-FQF model delivers comparable or superior performance compared to the DNN-based FQF model and significantly outperforms the conversion-based Spiking-FQF model across 19 Atari games.

\subsection{Comparing with other SNN-based DRL models}
Applying spiking neural networks to deep reinforcement learning tasks presents numerous challenges, and some studies have carried out preliminary explorations. For instance, \citep{tan2021strategy} utilized the ANN-SNN conversion method to transform the DNN model in the Deep Q Network into an SNN, achieving comparable scores on Atari games. Additionally, directly trained SNN-based models such as SDQN~\citep{liu2021human,chen2022deep} have been proposed, and the PL-SDQN model~\citep{sun2022solving} introduced membrane potential normalization to address the issue of spiking information vanishing during the training of deep SNNs in reinforcement learning tasks. We compared the performance of the proposed MCS-FQF model with these models on Atari game tasks, and the results, as shown in Table~\ref{Tab:sdqn_comp}, indicate that the MCS-FQF model outperformed the other models on most of the tasks.

The results in Table ~\ref{Tab:sdqn_comp} show that compared to the model based on ANN-SNN conversion method, the SNN deep learning model based on the STBP direct training method can achieve better performance. Compared to the Convert-SDQN model, both the SDQN models using the direct training method and our proposed MCS-FQF model were able to achieve better scores in most game scenarios. This paper introduces a direct training method for the spike-based FQF model of multi-compartment spiking neural networks based on temporal backpropagation,  and comparing it with other SDQN models under the condition of training on the same two million frames. 

The proposed MCS-FQF model outperformed the SDQN model using direct training methods on all 16 tested game tasks. Compared to the PL-SDQN model that utilizes membrane potential normalization, the MCS-FQF model demonstrated superior performance in 12 of these tasks. From the analysis of Table~\ref{Tab:game_res} and the results of the ablation experiments discussed later, it can be concluded that the performance improvement of the proposed model can be attributed to two main factors. On one hand, it benefits from integrating spiking neural networks with a more complex distributional reinforcement learning framework. On the other hand, the multi-compartment spiking neurons and their training method enhance the ability of the SNN model to integrate information in complex decision-making task environments.

\textbf{
	\begin{table*}[!]
		\centering
		\resizebox{1.0\textwidth}{!}{
			\begin{tabular}{l l l l  l l }	
				\toprule[2pt]
				& \textbf{Convert-SDQN}  & \textbf{SDQN}      & \textbf{SDQN}       &\textbf{PL-SDQN} & \textbf{MCS-FQF} \\
				& \textbf{(Tan2021)}    &\textbf{(Liu2021)}  & \textbf{(Chen2022)} &\textbf{(Sun2022)} & \textbf{(Ours)}  \\   
				\midrule[1pt]
				Atlantis             & 177034   & 487366.7    & 2481620.0  & \textbf{3267760.0}   & 3267690.0 \\
				BeamRider    		 & 9189.3   & 7226.9      & 5188.9     & 11480.4     & \textbf{15302.2}\\
				Boxing               & 75.5     & 95.3        & 84.4       & 99.5        & \textbf{99.8}\\ 
				Breakout     		 & 286.7    & 386.5       & 360.8      & 427.7       & \textbf{501.5} \\	
				CrazyClimber 		 & 106416   & 123916.7 	  & 93753.3    & 147950.0    & \textbf{163010.0}\\
				Gopher               & 6691.2   & 10107.3     & 4154.0     & \textbf{24064.0}     & 12812.0\\
				Jamesbond            & 521      & 1156.7      & 463.3      & 1460.0      & \textbf{1670.0}\\
				Kangaroo	  		 & 9760     & 8880.0      & 6140.0     & 14500.0     & \textbf{15080.0}\\
				Krull	      		 & 4128.2   & 9940.0      & 6899.0     & 11807.0     & \textbf{11876.0}\\
			    NameThisGame       & 8448     & 10877.0     & 7082.7     & \textbf{12202.0}     & 9268.0\\
			    Pong                 & 19.8     & 20.3        & 19.1       & 20.0        & \textbf{20.8}\\
				RoadRunner          & 41588    & 48983.3     & 23206.7    & 51930.0     & \textbf{582790.0}\\
				SpaceInvaders       & 2256.8   & 1832.2      & 1132.3     & 2433.5      & \textbf{2909.5}\\
				StarGunner          & 54692    & 57686.7     & 1716.7     & 63560.0     & \textbf{103100.0}\\
				Tutankham            & 157.1    & 194.7       & 276.0      & 271.5       & \textbf{298.5}\\
				VideoPinball        & 74012.1  & 275342.8    & 441615.2   & \textbf{673553.0}    & 606765.7\\ 
				\bottomrule[2pt]		
		\end{tabular}}
		\caption{Comparing different SNN-based deep reinforcement learning models on Atari games. The highest score for each game is bolded.}
		\label{Tab:sdqn_comp}
	\end{table*}
}

\subsection{Ablation study on the effect of MCN and population encoding}

\begin{figure}[!htb]
	\centering
	\includegraphics[width=1.0\linewidth]{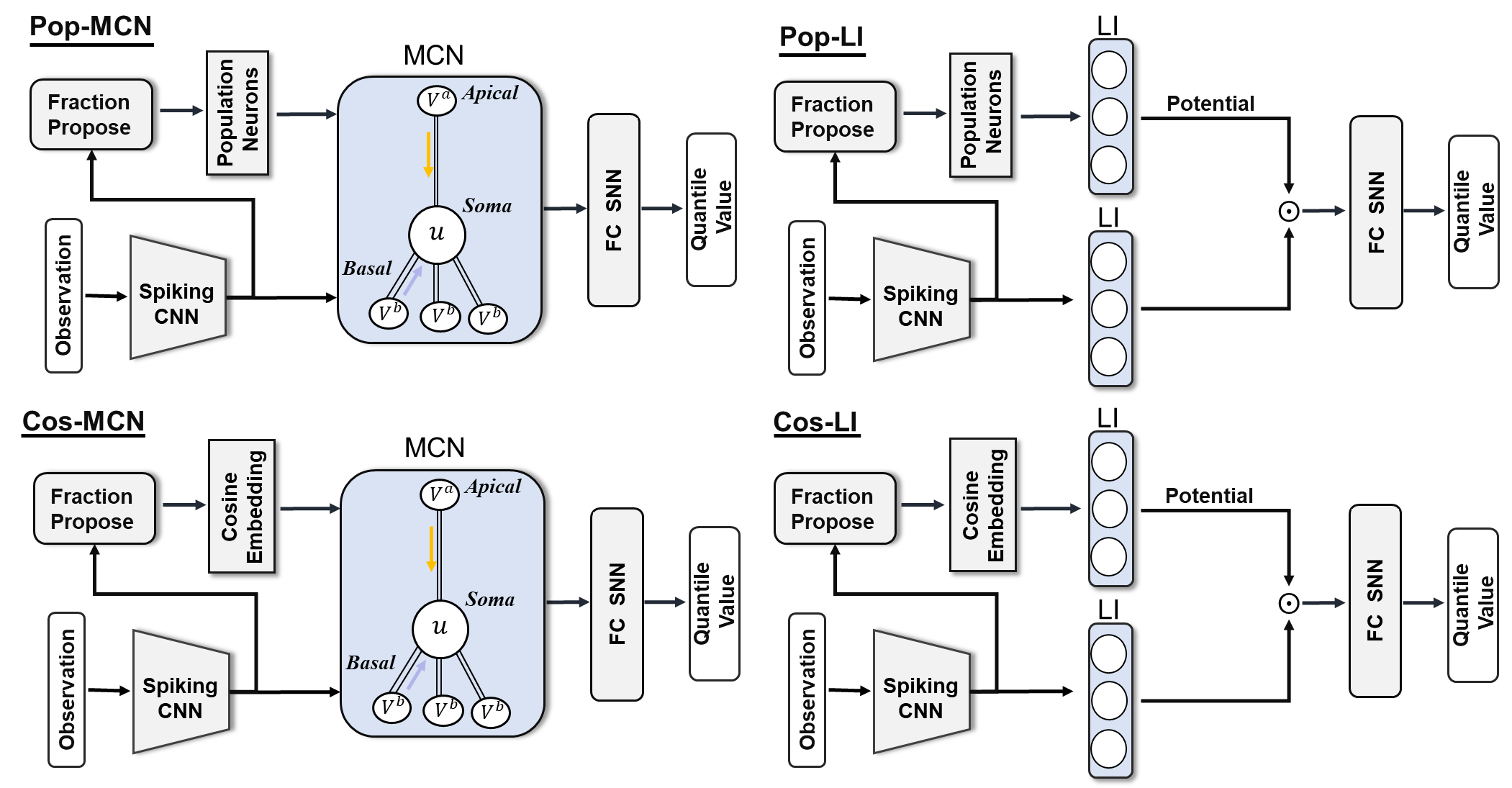} 
	\caption{Comparison models for ablation study. \textbf{Pop-MCN}: The proposed MCS-FQF model utilizing population embedding and the MCN model. \textbf{Cos-MCN}: Cosine embedding with the MCN model. \textbf{Pop-LI}: Population embedding with two groups of LI neurons for information integration. \textbf{Cos-LI}: Cosine embedding with LI neurons.}
	\label{Fig:ablation_model}
\end{figure}

To further analyze the respective contributions of MCN and population encoding in the proposed model, we perform the model ablation studie on the Atari environment.
Firstly, to demonstrate the advantages of the multi-compartment neuron model, a comparative experiment was conducted using two groups of Leaky-Integrate (LI) neurons instead of the MCN model to process state embedding spikes and quantile fraction embedding spikes respectively. In order to maintain consistency with the parameter count of the MCN model, each group consisted of 512 LI neurons. The LI neuron model is LIF model without the spiking process, which uses the accumulated membrane potential $u(t)$ to transmit information~\citep{orhan2012leaky}. Because the discrete spike signal outputs by the LIF model cannot directly represent continuous real values and perform signal integration operations, like adding or multiplying directly. The membrane potential output of LI neurons is widely used in applying SNN model on reinforcement learning tasks~\citep{chen2022deep, liu2021human}.  Each step's output membrane potentials $u(t)$ of these two groups of neurons are multiplied as the integration of state and fraction embedding. Then the integrated information is input to the final fully connected SNN to generate quantile values, as shown in Figure~\ref{Fig:ablation_model}. 
Then we examined the role of the population encoding  method in mapping continuous real values to the spike space, comparing it with the cosine coding method used by the FQF model. The quantile fraction $\tau$ generated from state embedding spikes are represented by $cos(i\pi\tau)$ with $i\in\{0, 1,...,M-1\}$.

\begin{figure*}[!htb]
	\centering
	\includegraphics[width=0.95\linewidth]{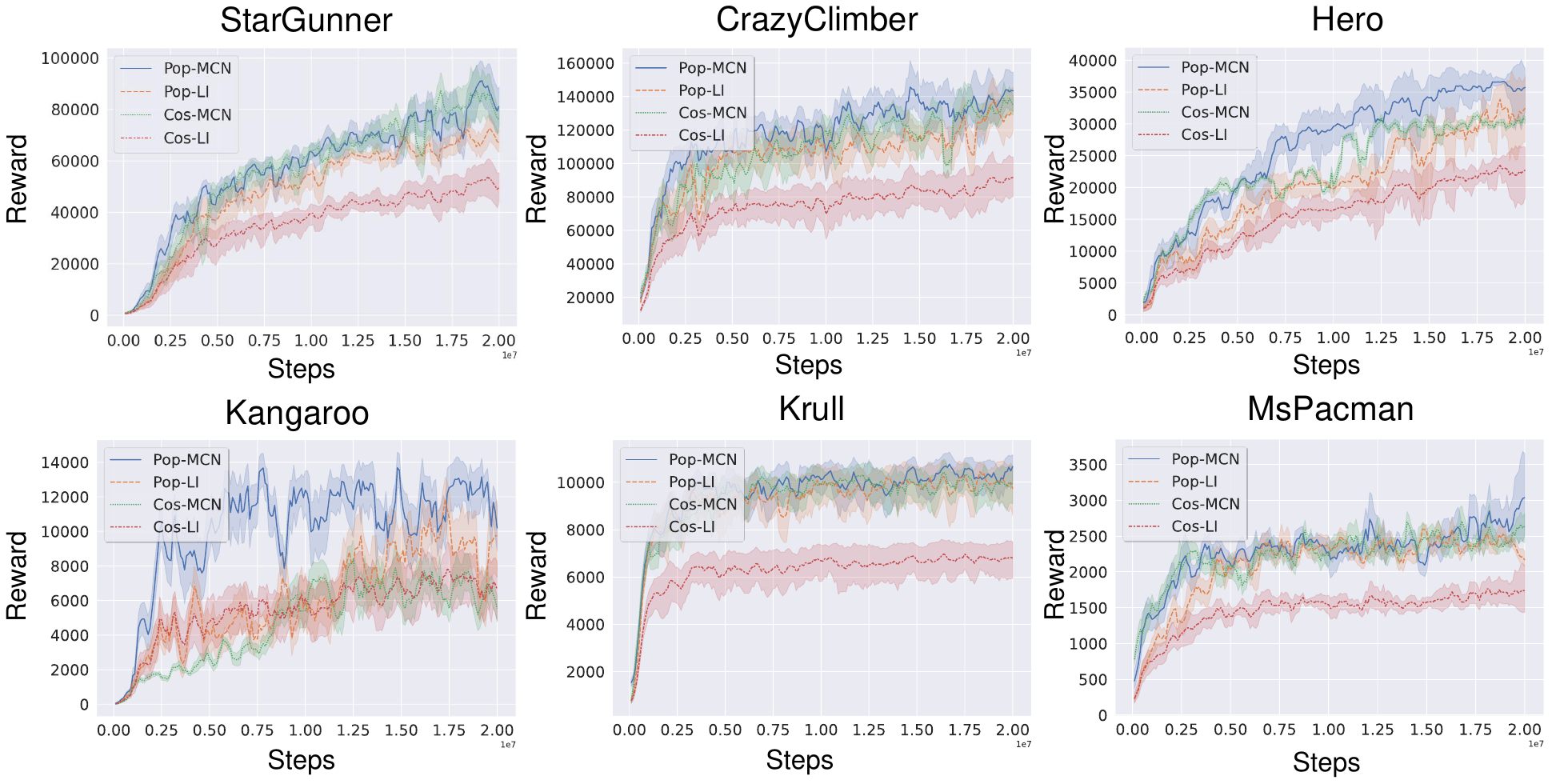} 
	\caption{The effects of MCN and population encoding on models' performance.}
	\label{Fig:abliation}
\end{figure*}

We train all three models with the same parameter settings and run ten rounds of testing experiments for each model. The results are shown in Figure~\ref{Fig:abliation} and Table~\ref{Tab:ablation}. Without MCN for information integration, the POP-LI model shows performance degradation compared with MCS-FQF.
The similar performance degradation is observed in the comparison between the Cos-LI and Cos-MCN models, both of which use the same fractional cosine embedding method as the FQF model to replace the population encoding method. 
Besides, the population encoding method also plays a critical part in the model. The Cos-LI model learns more slowly and gets much small average scores than POP-LI and MCS-FQF models. It is noteworthy that, under the same conditions of information integration using the MCN model, Cos-MCN, which employs cosine embedding, generally exhibits only a minor degradation in performance across most gaming tasks compared to Pop-MCN, which utilizing population encoding. This decline is smaller than the performance gap observed between Cos-LI and Pop-LI. This indirectly indicates that the information integration capability of the MCN model we proposed can adapt effectively to different fraction embedding methods. Additionally, the Pop-MCN model achieved the highest scores in most tasks, validating the effectiveness of the proposed MCN model and the Population embedding method.

\begin{table*}[!]
	\centering
	\resizebox{1.0\textwidth}{!}{
		\begin{tabular}{l l l l  l l l l l l l l l}	
			\toprule[2pt]
			& \multicolumn{3}{c}{\textbf{Cos-LI}} & \multicolumn{3}{c}{\textbf{Pop-LI}} & \multicolumn{3}{c}{\textbf{Cos-MCN}} &\multicolumn{2}{c}{\textbf{Pop-MCN(MCS-FQF)}} \\
			\textbf{Game} & Score & $\pm$std & (Score$\%$)  & Score & $\pm$std &(Score$\%$)  & Score & $\pm$std &(Score$\%$)  & Score & $\pm$std & (Score$\%$)   \\
			\midrule[1pt]
			Asteroids             &967.0   &153.1  &(15.83$\%$)              &1517.0      &406.5    & (26.80$\%$)        &1564.0 & 382 & (24.42$\%$)                 & \textbf{1666.0} & 718.4  &(43.12$\%$)      \\
			Atlantis              &1545819.1   &232356.4  &(15.03$\%$)       &2847970.0   &42491.8  &(1.49$\%$)          &3244965.7 & 74417.5 &  (2.30$\%$)          & \textbf{3267690.0} & 120276.6  &(3.68$\%$) \\
			BeamRider    		  &8291.4   &3400.6     &(41.01$\%$)         &12800.8     &3922.3   &(30.64$\%$)         &13055.4 & 6169 & (47.25$\%$)               & \textbf{15302.2}   & 6990.5   &(45.68$\%$) \\
			Bowling               &41.9   &2.8     &(6.67$\%$)               &\textbf{69.6}   &4.7     &(6.75$\%$)       &65.1 &5.4 & (8.29$\%$)                     & 63.5   & 8.5   &(13.46$\%$) \\
			Breakout     		  &280.9   &123.9     &(44.11$\%$)           &419.8   &119.3     &(28.41$\%$)            &493.1 &154.7 & (31.37$\%$)                 & \textbf{501.5}      & 170.3      &(33.96$\%$) \\
			Centipede    		  &6619.1   &2494.1     &(37.68$\%$)	     &\textbf{7979.1} &2932.0   &(36.75$\%$)     &6391.6 &2259.1 & (35.34$\%$)               & 7375.9    & 2929.6  &(39.72$\%$) \\
			CrazyClimber 		  &91873.3   &23451.3     &(25.53$\%$)  	 & 143020.0  &19430.7     &(13.59$\%$)       &148261.0 &28402.3 & (19.16$\%$)            & \textbf{163010.0}    & 21402.8  &(13.13$\%$) \\
			Enduro                &1574.7   &328.0     &(20.83$\%$)	         &2965.4   &1174.7   &(39.61$\%$)            &4133.2 &1325.6 & (32.07$\%$)               & \textbf{4156.4}    & 874.7  &(21.04$\%$) \\
			Frostbite             &3452.5    &646.7     & (18.73$\%$)        &4889.0   &894.6    &(18.30$\%$)            &8358.0  &898.1 & (10.74$\%$)               & \textbf{8733.0}   &933.6  &(10.69$\%$) \\
			Hero                  &23576.8   &348.6     &(1.48$\%$)	         &34071.0   &1390.9  &(4.08$\%$)             &32546.0 & 1423.2 & (4.37$\%$)              & \textbf{36912.5}    & 1091.4  &(2.96$\%$) \\
			Kangaroo	  		  &7111.6   &1439.8     &(20.25$\%$)         &13100.0   &2218.1     &(16.93$\%$)         &9235.2  & 1356.3 & (14.69$\%$)             & \textbf{15080.0}     & 325.0    &(2.15$\%$) \\
			Krull	      		  &9072.9   &3149.9     &(34.72$\%$)         &11114.0   &309.8     &(2.79$\%$)           &10722.0 & 1167.5 & (10.89$\%$)             & \textbf{11276.0}     & 599.5    &(5.32$\%$) \\
			KungFuMaster 		  &20466.7   &4120.2     &(20.13$\%$)        & \textbf{35770.0} &10032.2 &(28.05$\%$)    &35276.0 & 8175.4 & (23.18$\%$)             & 34820.0     & 9692.9    &(27.84$\%$) \\
			MsPacman	  		  &1650.5   &534.7   &(32.40$\%$)	         & 3017.0 &1055.8   &(34.99$\%$)             &2843.0 & 840.5  & (29.56$\%$)              & \textbf{3841.0}	   & 1269.7   &(33.06$\%$) \\
			Qbert                 &8818.2   &725.6   &(8.23$\%$)             & 16355.0  &1811.6 &(11.08$\%$)             &17723.1 & 1158.2 & (6.53$\%$)              & \textbf{18042.5}     & 1577.5   &(8.74$\%$)\\
			RoadRunner            &32518.2 &3813.8 &(11.73$\%$)              &50670.0 &11126.9 &(21.96$\%$)              &49523.0 &9729.3 & (19.64$\%$)              &\textbf{52790.0}         & 13469.5   &(25.52$\%$)\\
			StarGunner            &43440.7   &9515.0     &(21.90$\%$)        & 80350.0   &10540.8  &(13.12$\%$)          &98129.0 &16532.4 & (16.85$\%$)             & \textbf{103100.0}     & 17266.19  &(16.75$\%$) \\
			Tutankham             &104.6 	 &20.7  & (19.75$\%$)            &214.4   &27.3     &(12.71$\%$)             &259.2 & 34.35 & (13.25$\%$)                & \textbf{298.5}       & 37.0  &(12.41$\%$) \\
			VideoPinball 		  &173795.0 &170014.4 &(97.82$\%$)           & 506845.2 &226578.6 &(44.70$\%$)           & 406556.0 & 274981.2 & (67.64$\%$)         & \textbf{606765.7}    & 250377.9  &(41.26$\%$) \\ 
			\bottomrule[2pt]		
	\end{tabular}}
	\caption{Scores for ablation study. All models are run ten times with the same experimental settings.}
	\label{Tab:ablation}
\end{table*}

\subsection{Computational complexity and energy efficiency}
Compared to the two LI neural networks used for information integration, the multi-compartment neuron model reduces computational complexity and conserves resources. This is primarily because multiplication operations consume significantly more resources than addition operations. To reflect the complexity of model computations, we measure the number of multiplications required by the network. In artificial neural networks, neurons process inputs by multiplying a floating-point (FP) weight matrix with the input vector and then summing the results, which corresponds to MAC operations on a computing device. In contrast, spiking neural networks process binary spike information using only FP addition in event-driven neuromorphic hardware~\citep{merolla2014million,furber2014spinnaker,davies2018loihi}. When an input neuron fires a spike, the corresponding weight value of the connected neurons in the following layer is recorded. If the input neuron does not fire a spike, the weight value is simply ignored. Finally, the weight values associated with spike inputs are summed to generate the membrane potential.

We have compiled the parameter counts for all network layers of the model, including the convolutional layers for visual processing, the fraction encoding layers, the MCN and LI layers used for feature fusion, and the final fully connected (SNN) layer used for outputting quantile values. All compared model use direct pixel value input for the first encoding layer of  SCNN. The SCNN model has the same computational cost as CNN in the first encoding layer. Defining $\mathbb{N}_{MAC}$ for the number of MAC operations, $\mathbb{N}_{ADD}$ for the number of addition operations, $\bar{r}$  for the average neuron firing rate of the whore model layers, and $T$ is simulation period. The energy of SNN model in calculation operation is $\mathbb{E}_{snn}$ obtained with  Eq.~(\ref{EQ:snn_eng}).
\begin{equation}
	\mathbb{E}_{snn}=  a\times\mathbb{N}_{MAC} + \bar{r}\times b\times \mathbb{N}_{ADD}\times T
	\label{EQ:snn_eng}
\end{equation}
Similarly, we can also get the energy consumption of ANN model as:
\begin{equation}
	\mathbb{E}_{ann}= a\times\mathbb{N}_{MAC} + b\times \mathbb{N}_{ADD}
	\label{EQ:ann_eng}
\end{equation}

We use the same 45-nm CMOS energy benchmark as in~\citep{rathi2021diet}, where the unit energy cost for a MAC operation is $a=4.6pJ$ and for an ADD operation is $b=0.9pJ$. The energy consumption of different models is compared in Table~\ref{Tab:energy_cost} ("M" denotes million). The ANN-based FQF model~\citep{yang2019fully} utilizes ReLU neurons, which primarily perform MAC operations throughout the computation process. In contrast, the SNN model replaces some of these operations with addition operations. The results show that the multi-compartment neuron and spiking population encoding methods not only enhance the performance of the MCS-FQF model but also reduce its energy consumption. Additionally, compared to Pop-LI/Cos-LI, the proposed MCS-FQF model performs fewer MAC operations and maintains a more optimal firing rate, striking a better balance between performance and power consumption.

\begin{table}[t]
	\centering
	\resizebox{.95\columnwidth}{!}{
		\begin{tabular}{l l l l l l}
			\toprule[2pt]
			Architecture & $\mathbb{N}_{ADD}$& $\mathbb{N}_{MAC}$ & $\bar{r}$ & T & Normalized Energy \\ 
			& & & & & ($\mathbb{E}/\mathbb{E}_{MCS-FQF}$)  \\
			\midrule[1pt]
			FQF-ANN & - & 9.42M & - & - & 3.2\\
			Cos-LI & 7.01M & 4.26M & 0.141 & 8ms & 1.9   \\
			Pop-LI & 7.61M & 3.46M & 0.0268 & 8ms & 1.2   \\
			MCS-FQF & 8.62M & 2.45M & 0.0342 & 8ms & 1.0   \\
			\bottomrule[2pt]
	\end{tabular}}
	\caption{Energy consumption of different models.}
	\label{Tab:energy_cost}
\end{table}

\section{Discussion}
Compared with the baseline FQF model and the Spiking-FQF model based on ANN-SNN conversion, our proposed MCS-FQF model demonstrates clear performance advantages. Ablation experiments further confirm the significant impact of the multi-compartment neuron model and the population encoding method on the model's information processing capabilities. In this section, we analyze the spiking activity of the MCN during the learning process of the MCS-FQF model and explore the interaction between the basal and apical dendrites in information integration. Based on the experimental results, we outline the factors contributing to the superior performance of the MCS-FQF model. Additionally, we discuss the limitations of the current study and provide suggestions for future research directions.

\subsection{Analysis about the spiking activity of MCN }
We recorded the multi-compartment neuron spiking activities when the trained model was tested on the MsPacman game to show the effect of basal dendrite and apical dendrite on the process of neuron making decision. A total of 128 multi-compartmental neurons in the MCS-FQF model were randomly selected, and their spiking activity was recorded over the testing trajectory. 

\begin{figure*}[!]
	\centering
	\includegraphics[width=0.9\linewidth]{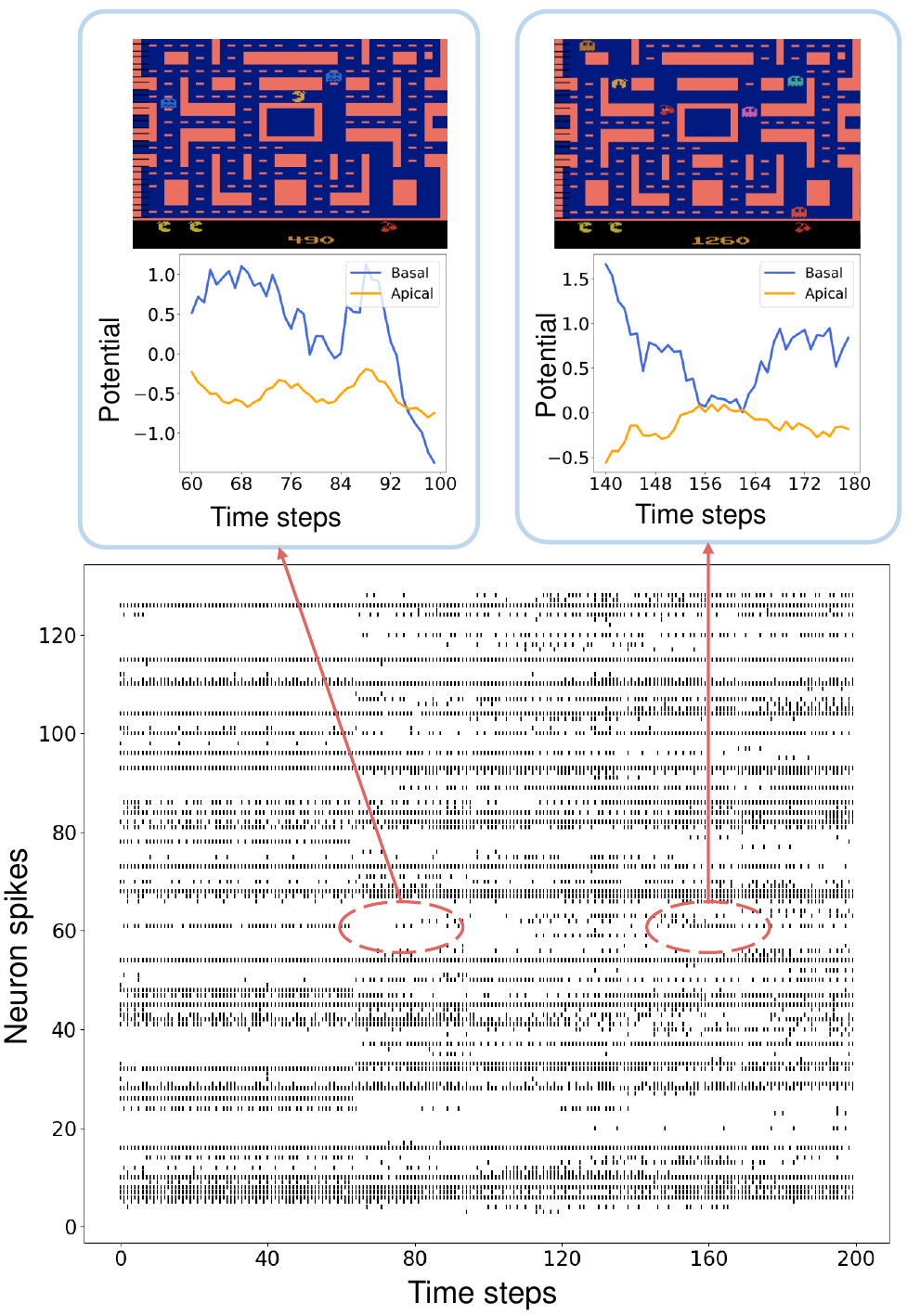} 
	\caption{Records of multi-compartment neuron spikes and dendritic potential on experiments.}
	\label{Fig:hist}
\end{figure*}

As shown in Figure~\ref{Fig:hist}, the multi-compartment neurons integrate state and quantile fraction information, resulting in distinct spike sequences. The two block diagrams at the top of Figure~\ref{Fig:hist} illustrate the changes in basal and apical dendritic potentials of the MCN. The diagram on the left demonstrates how apical dendrites inhibit the influence of basal dendrites on the somatic potential. When the basal dendritic potential is relatively high, the negative potential of the apical dendrite reduces the number of neuronal spikes.

In contrast, the diagram on the right of Figure~\ref{Fig:hist} shows that when the apical dendritic potential rises, even if the basal dendritic potential is low, the combined effects of both compartments can increase the intensity of neuronal spikes. We suggest that this mechanism of mutual inhibition and cooperation among different parts of the neuron enhances the computational power of a single neuron. SNNs incorporating multi-compartment neurons can thus benefit from more diverse and distinguishable neuronal spiking activities.

Ablation experiments reveal that the neuron structure plays a crucial role in the information integration process, with the MCN model outperforming both the combination of two groups of point neuron models and the pointwise value multiplication in FQF. We attribute this to the MCN's ability to integrate inputs from different sources at the spatiotemporal level, as described in Eq.~(\ref{EQ:mc_po}). Its dynamic accumulation and decay mechanisms are more effective than those of point neuron models in processing information represented by spike sequences. Moreover, compared to implicit cosine representations, encoding fractions with population spiking neurons transforms information into a high-dimensional space, making it more suitable for spiking neural networks to distinguish and process.

We analyzed the factors affecting the performance of the proposed MCS-FQF model:
\begin{itemize}
	\item Compared with the ReLU neurons in FQF, the spiking neuron model in MCS-FQF has more complex dynamic characteristics, which makes the model learn more information in the RL environment.
	\item Unlike pointwisely multiplying state embedding and fraction embedding in FQF and ANN-SNN conversion-based Spiking-FQF, MCS-FQF uses the multi-compartment neuron model to process them and has better information integrating capabilities.
	\item The directly trained MCS-FQF can learn better network weights than the Spiking-FQF model based on the ANN-SNN conversion.
	\item The population encoding method lets the quantile fraction information be more effectively represented in the spiking neural network.
\end{itemize}

\subsection{Analysis of MCN model in classification tasks}
To validate the effectiveness of the multi-compartment neuron (MCN) model in different types of tasks, we conducted additional experimental analyses on the MNIST, sequential MNIST (S-MNIST), and Spiking Heidelberg Digits (SHD) classification tasks. For the MNIST and SHD datasets, we utilized a three-layer fully connected network architecture, with the hidden layer composed of 256 MCN neurons. The comparison models included LIF  and LI neural networks with the same fully connected architecture. To maintain consistency with the comparisons made in the RL tasks, the LI neuron networks were configured to match the two neuron groups (LI-Com), Pop-LI and Cos-LI, used in the ablation experiments. Considering that MCN neurons receive inputs at both apical and basal dendritic compartments, the hidden layer of the LIF network was set to 512 neurons, while the LI network was configured in a 256+256 format, corresponding to the number of LI neurons in the two neuron groups.

For the S-MNIST experiments, we employed a 6-layer convolutional network architecture. To maintain parameter consistency when comparing the MCN model with the LIF/LI models, we doubled the number of feature channels in the convolutional layers of the LIF model, similar to the approach used for the hidden layers in the fully connected network. Detailed model configurations and experimental results are provided in Table~\ref{Tab:simple}.

\begin{table*}[htbp]
	\centering
		\begin{tabular}{c|ccccccccccc}
			\toprule
			Dataset  & Method  & Architecture  & Time Step & Hidden/Channels  & Accuracy(\%)\\
			\midrule
			\multirow{3}{*}{SHD}  
			&  LIF                 & FCNet         & 15  & 512     & 63.11 \\
			& LI-Com               & FCNet         & 15  & 256+256 & 48.80 \\
			& \textbf{MCN(Ours)}   & FCNet         & 15  & 256     & \textbf{89.39} \\
			\midrule
			\multirow{3}{*}{MNIST}  
			&  LIF                 & FCNet         & 8   & 512     & 96.80 \\
			& LI-Com               & FCNet         & 8   & 256+256 & 93.81 \\
			& \textbf{MCN(Ours)}   & FCNet         & 8   & 256     & \textbf{97.32} \\
			\midrule
			\multirow{3}{*}{S-MNIST}  
			&  LIF                 & ConvNet       & 28  & 256     & 90.07 \\
			& LI-Com               & ConvNet       & 28  & 128+128 & 84.36 \\
			& \textbf{MCN(Ours)}   & ConvNet       & 28  & 128     & \textbf{94.61} \\
			\bottomrule
		\end{tabular}
		\caption{Comparison of different neural models on classification datasets.}
		\label{Tab:simple}
	\end{table*}
	
The experiments conducted on visual and temporal data classification tasks further validate the superiority of the MCN model in handling spatiotemporal feature data. The MNIST image dataset, which only contains spatial dimension feature information, shows that the MCN model achieved better accuracy compared to the LIF and LI models, although the differences between the three were relatively  small. On the SHD temporal sequence data, the MCN model achieved an accuracy of 89.39$\%$, significantly higher than the 63.11$\%$ and 48.80 $\%$ achieved by the LIF and LI models, respectively, thereby demonstrating the MCN model's performance advantage in processing temporal dimension information. Moreover, on the S-MNIST dataset, which contains both temporal dimension information and spatial visual features, the MCN model also outperformed the LIF and LI neuron models. This result, together with the performance of the MCS-FQF model in RL tasks, collectively supports the MCN model's advantage in processing spatiotemporal feature information.

\subsection{Influence of hyperparameters on MCN models}
Compared to the LIF neuron model, the multi-compartmental neuron model incorporates more detailed representations of the biological structures of neurons, thereby offering greater biological plausibility. However, the inclusion of these additional biological details results in more hyperparameters in the MCN model compared to the LIF neuron. In this study, we examined the impact of different settings of dendritic time constants $\tau_A$ and $\tau_B$, apical dendritic conductance $g_A$, and basal dendritic conductance $g_B$ on the performance of the MCS-FQF model. The experimental results for various hyperparameter settings on the Krull and MsPacman tasks are presented in Table~\ref{Tab:hyper_exp}. 
For the dendritic time constants $\tau_A$ and $\tau_B$, we simultaneously varied both parameters in the experiments, assigning them the same value. According to Eq.~(\ref{EQ:mc_fuse}), the ratios $g_b/g_L$ and $g_A/g_L$ are critical factors directly influencing the dendritic function of the MCN model. Therefore, the experiments compared the effects of varying these ratios on the performance of the MCS-FQF model.
\begin{table}[!]
	\centering
	\resizebox{0.5\textwidth}{!}{
		\begin{tabular}{l l l l  l l l l l l l l  lllllll}	
			\toprule
			\textbf{$\tau_A$\&$\tau_B$}    &\textbf{1.5}   &\textbf{1.75}  &\textbf{2.0}  &\textbf{2.5}  &\textbf{2.7} \\
			\midrule
			Krull                          &10603.0        &10848.0        &11276.0       &11176.0           &11439.0 \\
			MsPacman                       &2995.0         &3114.0         &3841.0        &2787.0           &3115.0 \\
			\midrule
			\textbf{$g_B/gL$}              &\textbf{0.5}   &\textbf{0.75}  &\textbf{1.5}  &\textbf{1.75}     &\textbf{2.0} \\
			\midrule
			Krull                          &10794.0        &11028.0        &10500.0       &10874.0           &10762.0 \\
			MsPacman                       &3360.0         &3974.0         &3309.0        &2723.0            &3300.0 \\
			\midrule
			\textbf{$g_A/gL$}              &\textbf{0.5}   &\textbf{0.75}  &\textbf{1.5}  &\textbf{1.75}     &\textbf{2.0} \\
			\midrule
			Krull                          &10954.0        &11275.0        &10689.0       &10704.0           &10695.0 \\
			MsPacman                       &3159.0         &2781.0         &3625.0        &3728.0            &3423.0  \\
			\bottomrule		
	\end{tabular}}
	\caption{Results for different setting of MCN's hyperparameters.}
	\label{Tab:hyper_exp}
\end{table}

For the MCS-FQF model proposed in this work, most hyperparameters of the MCN model have an optimal range where the model performs best. Specifically, for the dendritic time constants  $\tau_A$ and $\tau_B$, setting them around 2.0 tends to yield the highest performance. Similarly, there are optimal ranges for $g_B/g_L$ and $g_A/g_L$. Given that the default settings in previous experiments were set to 1.0, and based on the results from Table~\ref{Tab:sdqn_comp} and Table~\ref{Tab:hyper_exp}, the estimated optimal ranges for $g_b/g_L$ and $g_A/g_L$ are 0.75–1.0 and 1.0–1.75, respectively. The differences between these ranges can be attributed to the fact that in the MCS-FQF model, apical dendrites receive fraction embedding while basal dendrites receive state observation embedding, with each playing a distinct role in the overall decision-making process. However, it should be noted that changes in the MCN hyperparameters have a relatively limited impact on the performance of the MCS-FQF model, with performance variations of around 30$\%$ in the MsPacman game and less than 10$\%$ in the Krull task.

\subsection{Limitations of the study}
When modeling the neuron structure, we only used the effect of dendritic potential on somatic potential, and does not include the characteristic of mutual interaction in dendrite and soma in biological neuron. Because, if containing mutual interaction of dendrite and soma in multi-compartment model, it introduces circle in computing graph, the MC neuron’s potential easily generate oscillation and exploding. Moreover, the backpropagation training method is not work for the computing circles. In future work, we will construct the MC neuron model including the somatic influence in dendrites and propose an efficient training method suiting to train SNNs with circle computing in more complex tasks. The proposed population neurons in this work has the same receptive field. How to use population neurons with different receptive fields and optimal activity to better represent spiking information is another problem that needs to be solved.
Furthermore, considering the impact of multi-compartmental neuron hyperparameters on model performance, a potential direction for future work is to explore automated methods for identifying optimal hyperparameter settings. This could include approaches such as grid search or heuristic algorithms based on genetic and evolutionary strategies.

\section{Conclusion}
In this work, we propose a multi-compartment neuron model inspired by the structure of biological neurons. This model uses dendrites to receive inputs from different sources and conducts the resulting dendritic potentials to the soma for integrated information processing. We apply multi-compartment spiking neural networks to deep distributional reinforcement learning, implementing an SNN-based FQF model. The proposed MCS-FQF model represents quantile fractions within the spiking information space using a Gaussian neuron population encoding method. Experimental results demonstrate that our model outperforms both the vanilla FQF and the ANN-SNN conversion-based Spiking-FQF across 19 Atari game tests. Additionally, ablation studies and energy analyses indicate that the proposed multi-compartment neuron and population encoding methods enhance the performance of spiking neural networks while reducing power consumption in distributional reinforcement learning. Compared to ANNs, brain-inspired spiking neural networks offer superior functionality in information processing with greater biological plausibility. In this paper, we introduce the structural details and information encoding characteristics of biological neurons into SNN-based reinforcement learning models, thereby enhancing the capability of spiking neural networks to handle complex learning tasks. Furthermore, we hope our work contributes to further advancements in neuromorphic computing model exploration.

\subsection*{Data and code availability}
The proposed Algorithm is developed using the BrainCog platform. The Python scripts can be downloaded from the GitHub repository:
\href{https://github.com/BrainCog-X/Brain-Cog/tree/main/examples/decision\_making/RL/mcs-fqf}{https://github.com/BrainCog-X/Brain-Cog/tre\-e/main/examples/decision\_making/RL/mcs-fqf}
\subsection*{Lead contact}
Further information and requests for resources and reagents should be directed to and will be fulfilled by the lead contact, Yi Zeng (yi.zeng@ia.ac.cn).

\section*{Acknowledgments}
This study was supported by the Postdoctoral Fellowship Program of CPSF (Grant No.GZC20232994), Special Research Assistant Program of the Chinese Academy of Sciences (Grant No.E4S9230501), the Funding from Institute of Automation, Chinese Academy of Sciences(Grant No. E411230101)

\subsection*{Declaration of interests}

The authors declare that they have no known competing financial interests or personal relationships that could have appeared to
influence the work reported in this paper.









\printcredits

\bibliographystyle{cas-model2-names}

\bibliography{Refs}

\begin{thebibliography}{53}
\expandafter\ifx\csname natexlab\endcsname\relax\def\natexlab#1{#1}\fi
\providecommand{\url}[1]{\texttt{#1}}
\providecommand{\href}[2]{#2}
\providecommand{\path}[1]{#1}
\providecommand{\DOIprefix}{doi:}
\providecommand{\ArXivprefix}{arXiv:}
\providecommand{\URLprefix}{URL: }
\providecommand{\Pubmedprefix}{pmid:}
\providecommand{\doi}[1]{\href{http://dx.doi.org/#1}{\path{#1}}}
\providecommand{\Pubmed}[1]{\href{pmid:#1}{\path{#1}}}
\providecommand{\bibinfo}[2]{#2}
\ifx\xfnm\relax \def\xfnm[#1]{\unskip,\space#1}\fi
\bibitem[{Balachandar and Michmizos(2020)}]{balachandar2020spiking}
\bibinfo{author}{Balachandar, P.}, \bibinfo{author}{Michmizos, K.P.},
  \bibinfo{year}{2020}.
\newblock \bibinfo{title}{A spiking neural network emulating the structure of
  the oculomotor system requires no learning to control a biomimetic robotic
  head}, in: \bibinfo{booktitle}{2020 8th IEEE RAS/EMBS International
  Conference for Biomedical Robotics and Biomechatronics (BioRob)},
  \bibinfo{organization}{IEEE}. pp. \bibinfo{pages}{1128--1133}.
\bibitem[{Capone et~al.(2022)Capone, Lupo, Muratore and
  Paolucci}]{capone2022burst}
\bibinfo{author}{Capone, C.}, \bibinfo{author}{Lupo, C.},
  \bibinfo{author}{Muratore, P.}, \bibinfo{author}{Paolucci, P.S.},
  \bibinfo{year}{2022}.
\newblock \bibinfo{title}{Burst-dependent plasticity and dendritic
  amplification support target-based learning and hierarchical imitation
  learning}.
\newblock \bibinfo{journal}{arXiv preprint arXiv:2201.11717} .
\bibitem[{Chen et~al.(2022)Chen, Peng, Huang and Tian}]{chen2022deep}
\bibinfo{author}{Chen, D.}, \bibinfo{author}{Peng, P.}, \bibinfo{author}{Huang,
  T.}, \bibinfo{author}{Tian, Y.}, \bibinfo{year}{2022}.
\newblock \bibinfo{title}{Deep reinforcement learning with spiking q-learning}.
\newblock \bibinfo{journal}{arXiv preprint arXiv:2201.09754} .
\bibitem[{Davies et~al.(2018)Davies, Srinivasa, Lin, Chinya, Cao, Choday,
  Dimou, Joshi, Imam, Jain et~al.}]{davies2018loihi}
\bibinfo{author}{Davies, M.}, \bibinfo{author}{Srinivasa, N.},
  \bibinfo{author}{Lin, T.H.}, \bibinfo{author}{Chinya, G.},
  \bibinfo{author}{Cao, Y.}, \bibinfo{author}{Choday, S.H.},
  \bibinfo{author}{Dimou, G.}, \bibinfo{author}{Joshi, P.},
  \bibinfo{author}{Imam, N.}, \bibinfo{author}{Jain, S.}, et~al.,
  \bibinfo{year}{2018}.
\newblock \bibinfo{title}{Loihi: A neuromorphic manycore processor with on-chip
  learning}.
\newblock \bibinfo{journal}{Ieee Micro} \bibinfo{volume}{38},
  \bibinfo{pages}{82--99}.
\bibitem[{Dominguez-Morales et~al.(2018)Dominguez-Morales, Liu, James,
  Gutierrez-Galan, Jimenez-Fernandez, Davidson and Furber}]{dominguez2018deep}
\bibinfo{author}{Dominguez-Morales, J.P.}, \bibinfo{author}{Liu, Q.},
  \bibinfo{author}{James, R.}, \bibinfo{author}{Gutierrez-Galan, D.},
  \bibinfo{author}{Jimenez-Fernandez, A.}, \bibinfo{author}{Davidson, S.},
  \bibinfo{author}{Furber, S.}, \bibinfo{year}{2018}.
\newblock \bibinfo{title}{Deep spiking neural network model for time-variant
  signals classification: a real-time speech recognition approach}, in:
  \bibinfo{booktitle}{2018 International Joint Conference on Neural Networks
  (IJCNN)}, \bibinfo{organization}{IEEE}. pp. \bibinfo{pages}{1--8}.
\bibitem[{Furber et~al.(2014)Furber, Galluppi, Temple and
  Plana}]{furber2014spinnaker}
\bibinfo{author}{Furber, S.B.}, \bibinfo{author}{Galluppi, F.},
  \bibinfo{author}{Temple, S.}, \bibinfo{author}{Plana, L.A.},
  \bibinfo{year}{2014}.
\newblock \bibinfo{title}{The spinnaker project}.
\newblock \bibinfo{journal}{Proceedings of the IEEE} \bibinfo{volume}{102},
  \bibinfo{pages}{652--665}.
\bibitem[{Gerstner and Kistler(2002)}]{gerstner2002spiking}
\bibinfo{author}{Gerstner, W.}, \bibinfo{author}{Kistler, W.M.},
  \bibinfo{year}{2002}.
\newblock \bibinfo{title}{Spiking neuron models: Single neurons, populations,
  plasticity}.
\newblock \bibinfo{publisher}{Cambridge university press}.
\bibitem[{Gidon et~al.(2020)Gidon, Zolnik, Fidzinski, Bolduan, Papoutsi,
  Poirazi, Holtkamp, Vida and Larkum}]{gidon2020dendritic}
\bibinfo{author}{Gidon, A.}, \bibinfo{author}{Zolnik, T.A.},
  \bibinfo{author}{Fidzinski, P.}, \bibinfo{author}{Bolduan, F.},
  \bibinfo{author}{Papoutsi, A.}, \bibinfo{author}{Poirazi, P.},
  \bibinfo{author}{Holtkamp, M.}, \bibinfo{author}{Vida, I.},
  \bibinfo{author}{Larkum, M.E.}, \bibinfo{year}{2020}.
\newblock \bibinfo{title}{Dendritic action potentials and computation in human
  layer 2/3 cortical neurons}.
\newblock \bibinfo{journal}{Science} \bibinfo{volume}{367},
  \bibinfo{pages}{83--87}.
\bibitem[{Hodgkin et~al.(1952)Hodgkin, Huxley and
  Katz}]{hodgkin1952measurement}
\bibinfo{author}{Hodgkin, A.L.}, \bibinfo{author}{Huxley, A.F.},
  \bibinfo{author}{Katz, B.}, \bibinfo{year}{1952}.
\newblock \bibinfo{title}{Measurement of current-voltage relations in the
  membrane of the giant axon of loligo}.
\newblock \bibinfo{journal}{The Journal of physiology} \bibinfo{volume}{116},
  \bibinfo{pages}{424}.
\bibitem[{Huber(1992)}]{huber1992robust}
\bibinfo{author}{Huber, P.J.}, \bibinfo{year}{1992}.
\newblock \bibinfo{title}{Robust estimation of a location parameter}, in:
  \bibinfo{booktitle}{Breakthroughs in statistics}.
  \bibinfo{publisher}{Springer}, pp. \bibinfo{pages}{492--518}.
\bibitem[{Izhikevich(2004)}]{izhikevich2004model}
\bibinfo{author}{Izhikevich, E.M.}, \bibinfo{year}{2004}.
\newblock \bibinfo{title}{Which model to use for cortical spiking neurons?}
\newblock \bibinfo{journal}{IEEE transactions on neural networks}
  \bibinfo{volume}{15}, \bibinfo{pages}{1063--1070}.
\bibitem[{Kampa and Stuart(2006)}]{kampa2006calcium}
\bibinfo{author}{Kampa, B.M.}, \bibinfo{author}{Stuart, G.J.},
  \bibinfo{year}{2006}.
\newblock \bibinfo{title}{Calcium spikes in basal dendrites of layer 5
  pyramidal neurons during action potential bursts}.
\newblock \bibinfo{journal}{Journal of Neuroscience} \bibinfo{volume}{26},
  \bibinfo{pages}{7424--7432}.
\bibitem[{Kim et~al.(2020)Kim, Park, Na and Yoon}]{kim2020spiking}
\bibinfo{author}{Kim, S.}, \bibinfo{author}{Park, S.}, \bibinfo{author}{Na,
  B.}, \bibinfo{author}{Yoon, S.}, \bibinfo{year}{2020}.
\newblock \bibinfo{title}{Spiking-yolo: spiking neural network for
  energy-efficient object detection}, in: \bibinfo{booktitle}{Proceedings of
  the AAAI conference on artificial intelligence}, pp.
  \bibinfo{pages}{11270--11277}.
\bibitem[{Kopsick et~al.(2022)Kopsick, Tecuatl, Moradi, Attili, Kashyap, Xing,
  Chen, Krichmar and Ascoli}]{kopsick2022robust}
\bibinfo{author}{Kopsick, J.D.}, \bibinfo{author}{Tecuatl, C.},
  \bibinfo{author}{Moradi, K.}, \bibinfo{author}{Attili, S.M.},
  \bibinfo{author}{Kashyap, H.J.}, \bibinfo{author}{Xing, J.},
  \bibinfo{author}{Chen, K.}, \bibinfo{author}{Krichmar, J.L.},
  \bibinfo{author}{Ascoli, G.A.}, \bibinfo{year}{2022}.
\newblock \bibinfo{title}{Robust resting-state dynamics in a large-scale
  spiking neural network model of area ca3 in the mouse hippocampus}.
\newblock \bibinfo{journal}{Cognitive Computation} , \bibinfo{pages}{1--21}.
\bibitem[{Lansdell et~al.(2019)Lansdell, Prakash and
  Kording}]{lansdell2019learning}
\bibinfo{author}{Lansdell, B.J.}, \bibinfo{author}{Prakash, P.R.},
  \bibinfo{author}{Kording, K.P.}, \bibinfo{year}{2019}.
\newblock \bibinfo{title}{Learning to solve the credit assignment problem}.
\newblock \bibinfo{journal}{arXiv preprint arXiv:1906.00889} .
\bibitem[{Lee et~al.(2020)Lee, Sarwar, Panda, Srinivasan and
  Roy}]{lee2020enabling}
\bibinfo{author}{Lee, C.}, \bibinfo{author}{Sarwar, S.S.},
  \bibinfo{author}{Panda, P.}, \bibinfo{author}{Srinivasan, G.},
  \bibinfo{author}{Roy, K.}, \bibinfo{year}{2020}.
\newblock \bibinfo{title}{Enabling spike-based backpropagation for training
  deep neural network architectures}.
\newblock \bibinfo{journal}{Frontiers in neuroscience} , \bibinfo{pages}{119}.
\bibitem[{Li et~al.(2023)Li, Tang and Lai}]{li2023learning}
\bibinfo{author}{Li, X.}, \bibinfo{author}{Tang, J.}, \bibinfo{author}{Lai,
  J.}, \bibinfo{year}{2023}.
\newblock \bibinfo{title}{Learning high-performance spiking neural networks
  with multi-compartment spiking neurons}, in:
  \bibinfo{booktitle}{International Conference on Image and Graphics},
  \bibinfo{organization}{Springer}. pp. \bibinfo{pages}{91--102}.
\bibitem[{Li and Zeng(2022)}]{li2022efficient}
\bibinfo{author}{Li, Y.}, \bibinfo{author}{Zeng, Y.}, \bibinfo{year}{2022}.
\newblock \bibinfo{title}{Efficient and accurate conversion of spiking neural
  network with burst spikes}.
\newblock \bibinfo{journal}{arXiv preprint arXiv:2204.13271} .
\bibitem[{Liu et~al.(2021)Liu, Deng, Xie, Huang and Tang}]{liu2021human}
\bibinfo{author}{Liu, G.}, \bibinfo{author}{Deng, W.}, \bibinfo{author}{Xie,
  X.}, \bibinfo{author}{Huang, L.}, \bibinfo{author}{Tang, H.},
  \bibinfo{year}{2021}.
\newblock \bibinfo{title}{Human-level control through directly-trained deep
  spiking q-networks}.
\newblock \bibinfo{journal}{arXiv preprint arXiv:2201.07211} .
\bibitem[{Lowet et~al.(2020)Lowet, Zheng, Matias, Drugowitsch and
  Uchida}]{lowet2020distributional}
\bibinfo{author}{Lowet, A.S.}, \bibinfo{author}{Zheng, Q.},
  \bibinfo{author}{Matias, S.}, \bibinfo{author}{Drugowitsch, J.},
  \bibinfo{author}{Uchida, N.}, \bibinfo{year}{2020}.
\newblock \bibinfo{title}{Distributional reinforcement learning in the brain}.
\newblock \bibinfo{journal}{Trends in Neurosciences} \bibinfo{volume}{43},
  \bibinfo{pages}{980--997}.
\bibitem[{Luo et~al.(2021)Luo, Xu, Yuan, Cao, Zhang, Xu, Wang and
  Feng}]{luo2021siamsnn}
\bibinfo{author}{Luo, Y.}, \bibinfo{author}{Xu, M.}, \bibinfo{author}{Yuan,
  C.}, \bibinfo{author}{Cao, X.}, \bibinfo{author}{Zhang, L.},
  \bibinfo{author}{Xu, Y.}, \bibinfo{author}{Wang, T.}, \bibinfo{author}{Feng,
  Q.}, \bibinfo{year}{2021}.
\newblock \bibinfo{title}{Siamsnn: siamese spiking neural networks for
  energy-efficient object tracking}, in: \bibinfo{booktitle}{International
  Conference on Artificial Neural Networks}, \bibinfo{organization}{Springer}.
  pp. \bibinfo{pages}{182--194}.
\bibitem[{Makara and Magee(2013)}]{makara2013variable}
\bibinfo{author}{Makara, J.K.}, \bibinfo{author}{Magee, J.C.},
  \bibinfo{year}{2013}.
\newblock \bibinfo{title}{Variable dendritic integration in hippocampal ca3
  pyramidal neurons}.
\newblock \bibinfo{journal}{Neuron} \bibinfo{volume}{80},
  \bibinfo{pages}{1438--1450}.
\bibitem[{Merolla et~al.(2014)Merolla, Arthur, Alvarez-Icaza, Cassidy, Sawada,
  Akopyan, Jackson, Imam, Guo, Nakamura et~al.}]{merolla2014million}
\bibinfo{author}{Merolla, P.A.}, \bibinfo{author}{Arthur, J.V.},
  \bibinfo{author}{Alvarez-Icaza, R.}, \bibinfo{author}{Cassidy, A.S.},
  \bibinfo{author}{Sawada, J.}, \bibinfo{author}{Akopyan, F.},
  \bibinfo{author}{Jackson, B.L.}, \bibinfo{author}{Imam, N.},
  \bibinfo{author}{Guo, C.}, \bibinfo{author}{Nakamura, Y.}, et~al.,
  \bibinfo{year}{2014}.
\newblock \bibinfo{title}{A million spiking-neuron integrated circuit with a
  scalable communication network and interface}.
\newblock \bibinfo{journal}{Science} \bibinfo{volume}{345},
  \bibinfo{pages}{668--673}.
\bibitem[{Meyers(2018)}]{meyers2018dynamic}
\bibinfo{author}{Meyers, E.M.}, \bibinfo{year}{2018}.
\newblock \bibinfo{title}{Dynamic population coding and its relationship to
  working memory}.
\newblock \bibinfo{journal}{Journal of neurophysiology} \bibinfo{volume}{120},
  \bibinfo{pages}{2260--2268}.
\bibitem[{Orhan(2012)}]{orhan2012leaky}
\bibinfo{author}{Orhan, E.}, \bibinfo{year}{2012}.
\newblock \bibinfo{title}{The leaky integrate-and-fire neuron model}.
\newblock \bibinfo{journal}{no} \bibinfo{volume}{3}, \bibinfo{pages}{1--6}.
\bibitem[{Panzeri et~al.(2015)Panzeri, Macke, Gross and
  Kayser}]{PANZERI2015162}
\bibinfo{author}{Panzeri, S.}, \bibinfo{author}{Macke, J.H.},
  \bibinfo{author}{Gross, J.}, \bibinfo{author}{Kayser, C.},
  \bibinfo{year}{2015}.
\newblock \bibinfo{title}{Neural population coding: combining insights from
  microscopic and mass signals}.
\newblock \bibinfo{journal}{Trends in Cognitive Sciences} \bibinfo{volume}{19},
  \bibinfo{pages}{162--172}.
\newblock \URLprefix
  \url{https://www.sciencedirect.com/science/article/pii/S1364661315000030},
  \DOIprefix\doi{https://doi.org/10.1016/j.tics.2015.01.002}.
\bibitem[{Poirazi et~al.(2003)Poirazi, Brannon and Mel}]{poirazi2003pyramidal}
\bibinfo{author}{Poirazi, P.}, \bibinfo{author}{Brannon, T.},
  \bibinfo{author}{Mel, B.W.}, \bibinfo{year}{2003}.
\newblock \bibinfo{title}{Pyramidal neuron as two-layer neural network}.
\newblock \bibinfo{journal}{Neuron} \bibinfo{volume}{37},
  \bibinfo{pages}{989--999}.
\bibitem[{Polsky et~al.(2004)Polsky, Mel and
  Schiller}]{polsky2004computational}
\bibinfo{author}{Polsky, A.}, \bibinfo{author}{Mel, B.W.},
  \bibinfo{author}{Schiller, J.}, \bibinfo{year}{2004}.
\newblock \bibinfo{title}{Computational subunits in thin dendrites of pyramidal
  cells}.
\newblock \bibinfo{journal}{Nature neuroscience} \bibinfo{volume}{7},
  \bibinfo{pages}{621--627}.
\bibitem[{Ponghiran and Roy(2022)}]{ponghiran2022spiking}
\bibinfo{author}{Ponghiran, W.}, \bibinfo{author}{Roy, K.},
  \bibinfo{year}{2022}.
\newblock \bibinfo{title}{Spiking neural networks with improved inherent
  recurrence dynamics for sequential learning}, in:
  \bibinfo{booktitle}{Proceedings of the AAAI Conference on Artificial
  Intelligence}, pp. \bibinfo{pages}{8001--8008}.
\bibitem[{Rathi and Roy(2021)}]{rathi2021diet}
\bibinfo{author}{Rathi, N.}, \bibinfo{author}{Roy, K.}, \bibinfo{year}{2021}.
\newblock \bibinfo{title}{Diet-snn: A low-latency spiking neural network with
  direct input encoding and leakage and threshold optimization}.
\newblock \bibinfo{journal}{IEEE Transactions on Neural Networks and Learning
  Systems} .
\bibitem[{Richards et~al.(2019)Richards, Lillicrap, Beaudoin, Bengio, Bogacz,
  Christensen, Clopath, Costa, de~Berker, Ganguli et~al.}]{richards2019deep}
\bibinfo{author}{Richards, B.A.}, \bibinfo{author}{Lillicrap, T.P.},
  \bibinfo{author}{Beaudoin, P.}, \bibinfo{author}{Bengio, Y.},
  \bibinfo{author}{Bogacz, R.}, \bibinfo{author}{Christensen, A.},
  \bibinfo{author}{Clopath, C.}, \bibinfo{author}{Costa, R.P.},
  \bibinfo{author}{de~Berker, A.}, \bibinfo{author}{Ganguli, S.}, et~al.,
  \bibinfo{year}{2019}.
\newblock \bibinfo{title}{A deep learning framework for neuroscience}.
\newblock \bibinfo{journal}{Nature neuroscience} \bibinfo{volume}{22},
  \bibinfo{pages}{1761--1770}.
\bibitem[{Sacramento et~al.(2018)Sacramento, Ponte~Costa, Bengio and
  Senn}]{sacramento2018dendritic}
\bibinfo{author}{Sacramento, J.}, \bibinfo{author}{Ponte~Costa, R.},
  \bibinfo{author}{Bengio, Y.}, \bibinfo{author}{Senn, W.},
  \bibinfo{year}{2018}.
\newblock \bibinfo{title}{Dendritic cortical microcircuits approximate the
  backpropagation algorithm}.
\newblock \bibinfo{journal}{Advances in neural information processing systems}
  \bibinfo{volume}{31}.
\bibitem[{Sanger(2003)}]{SANGER2003238}
\bibinfo{author}{Sanger, T.D.}, \bibinfo{year}{2003}.
\newblock \bibinfo{title}{Neural population codes}.
\newblock \bibinfo{journal}{Current Opinion in Neurobiology}
  \bibinfo{volume}{13}, \bibinfo{pages}{238--249}.
\newblock \URLprefix
  \url{https://www.sciencedirect.com/science/article/pii/S0959438803000345},
  \DOIprefix\doi{https://doi.org/10.1016/S0959-4388(03)00034-5}.
\bibitem[{Shan et~al.(2024)Shan, Zhang, Zhu, Qiu, Eshraghian and
  Qu}]{shan2024advancing}
\bibinfo{author}{Shan, Y.}, \bibinfo{author}{Zhang, M.}, \bibinfo{author}{Zhu,
  R.j.}, \bibinfo{author}{Qiu, X.}, \bibinfo{author}{Eshraghian, J.K.},
  \bibinfo{author}{Qu, H.}, \bibinfo{year}{2024}.
\newblock \bibinfo{title}{Advancing spiking neural networks towards multiscale
  spatiotemporal interaction learning}.
\newblock \bibinfo{journal}{arXiv preprint arXiv:2405.13672} .
\bibitem[{Shrestha et~al.(2021)Shrestha, Fang, Rider, Mei and
  Qiu}]{Shrestha2021In}
\bibinfo{author}{Shrestha, A.}, \bibinfo{author}{Fang, H.},
  \bibinfo{author}{Rider, D.P.}, \bibinfo{author}{Mei, Z.},
  \bibinfo{author}{Qiu, Q.}, \bibinfo{year}{2021}.
\newblock \bibinfo{title}{In-hardware learning of multilayer spiking neural
  networks on a neuromorphic processor}, in: \bibinfo{booktitle}{2021 58th
  ACM/IEEE Design Automation Conference (DAC)}, pp. \bibinfo{pages}{367--372}.
\newblock \DOIprefix\doi{10.1109/DAC18074.2021.9586323}.
\bibitem[{Smith et~al.(2013)Smith, Smith, Branco and
  H{\"a}usser}]{smith2013dendritic}
\bibinfo{author}{Smith, S.L.}, \bibinfo{author}{Smith, I.T.},
  \bibinfo{author}{Branco, T.}, \bibinfo{author}{H{\"a}usser, M.},
  \bibinfo{year}{2013}.
\newblock \bibinfo{title}{Dendritic spikes enhance stimulus selectivity in
  cortical neurons in vivo}.
\newblock \bibinfo{journal}{Nature} \bibinfo{volume}{503},
  \bibinfo{pages}{115--120}.
\bibitem[{Sun et~al.(2022)Sun, Zeng and Li}]{sun2022solving}
\bibinfo{author}{Sun, Y.}, \bibinfo{author}{Zeng, Y.}, \bibinfo{author}{Li,
  Y.}, \bibinfo{year}{2022}.
\newblock \bibinfo{title}{Solving the spike feature information vanishing
  problem in spiking deep q network with potential based normalization}.
\newblock \bibinfo{journal}{Frontiers in Neuroscience} \bibinfo{volume}{16}.
\newblock \URLprefix
  \url{https://www.frontiersin.org/articles/10.3389/fnins.2022.953368},
  \DOIprefix\doi{10.3389/fnins.2022.953368}.
\bibitem[{Tan et~al.(2021)Tan, Patel and Kozma}]{tan2021strategy}
\bibinfo{author}{Tan, W.}, \bibinfo{author}{Patel, D.}, \bibinfo{author}{Kozma,
  R.}, \bibinfo{year}{2021}.
\newblock \bibinfo{title}{Strategy and benchmark for converting deep q-networks
  to event-driven spiking neural networks}, in: \bibinfo{booktitle}{Proceedings
  of the AAAI conference on artificial intelligence}, pp.
  \bibinfo{pages}{9816--9824}.
\bibitem[{Tang et~al.(2020a)Tang, Kumar and Michmizos}]{tang2020reinforcement}
\bibinfo{author}{Tang, G.}, \bibinfo{author}{Kumar, N.},
  \bibinfo{author}{Michmizos, K.P.}, \bibinfo{year}{2020}a.
\newblock \bibinfo{title}{Reinforcement co-learning of deep and spiking neural
  networks for energy-efficient mapless navigation with neuromorphic hardware},
  in: \bibinfo{booktitle}{2020 IEEE/RSJ International Conference on Intelligent
  Robots and Systems (IROS)}, \bibinfo{organization}{IEEE}. pp.
  \bibinfo{pages}{6090--6097}.
\bibitem[{Tang et~al.(2020b)Tang, Kumar, Yoo and Michmizos}]{tang2020deep}
\bibinfo{author}{Tang, G.}, \bibinfo{author}{Kumar, N.}, \bibinfo{author}{Yoo,
  R.}, \bibinfo{author}{Michmizos, K.P.}, \bibinfo{year}{2020}b.
\newblock \bibinfo{title}{Deep reinforcement learning with population-coded
  spiking neural network for continuous control}.
\newblock \bibinfo{journal}{arXiv preprint arXiv:2010.09635} .
\bibitem[{Urbanczik and Senn(2014)}]{urbanczik2014learning}
\bibinfo{author}{Urbanczik, R.}, \bibinfo{author}{Senn, W.},
  \bibinfo{year}{2014}.
\newblock \bibinfo{title}{Learning by the dendritic prediction of somatic
  spiking}.
\newblock \bibinfo{journal}{Neuron} \bibinfo{volume}{81},
  \bibinfo{pages}{521--528}.
\bibitem[{Vaila(2021)}]{vaila2021deep}
\bibinfo{author}{Vaila, R.}, \bibinfo{year}{2021}.
\newblock \bibinfo{title}{Deep convolutional spiking neural networks for image
  classification}.
\newblock \bibinfo{publisher}{Boise State University}.
\bibitem[{Wang et~al.(2024a)Wang, Pan, Zhang, Tan and Li}]{wang2024restoring}
\bibinfo{author}{Wang, J.}, \bibinfo{author}{Pan, Z.}, \bibinfo{author}{Zhang,
  M.}, \bibinfo{author}{Tan, R.T.}, \bibinfo{author}{Li, H.},
  \bibinfo{year}{2024}a.
\newblock \bibinfo{title}{Restoring speaking lips from occlusion for
  audio-visual speech recognition}, in: \bibinfo{booktitle}{Proceedings of the
  AAAI Conference on Artificial Intelligence}, pp.
  \bibinfo{pages}{19144--19152}.
\bibitem[{Wang et~al.(2024b)Wang, Liu, Zhang, Luo and Qu}]{wang2024universal}
\bibinfo{author}{Wang, Y.}, \bibinfo{author}{Liu, H.}, \bibinfo{author}{Zhang,
  M.}, \bibinfo{author}{Luo, X.}, \bibinfo{author}{Qu, H.},
  \bibinfo{year}{2024}b.
\newblock \bibinfo{title}{A universal ann-to-snn framework for achieving high
  accuracy and low latency deep spiking neural networks}.
\newblock \bibinfo{journal}{Neural Networks} \bibinfo{volume}{174},
  \bibinfo{pages}{106244}.
\bibitem[{Wu et~al.(2018)Wu, Deng, Li, Zhu and Shi}]{Wu2018Spatio}
\bibinfo{author}{Wu, Y.}, \bibinfo{author}{Deng, L.}, \bibinfo{author}{Li, G.},
  \bibinfo{author}{Zhu, J.}, \bibinfo{author}{Shi, L.}, \bibinfo{year}{2018}.
\newblock \bibinfo{title}{Spatio-temporal backpropagation for training
  high-performance spiking neural networks}.
\newblock \bibinfo{journal}{Frontiers in Neuroscience} \bibinfo{volume}{12}.
\newblock \URLprefix
  \url{https://www.frontiersin.org/article/10.3389/fnins.2018.00331},
  \DOIprefix\doi{10.3389/fnins.2018.00331}.
\bibitem[{Yang et~al.(2019a)Yang, Zhao, Lin, Qin, Bian and Liu}]{yang2019fully}
\bibinfo{author}{Yang, D.}, \bibinfo{author}{Zhao, L.}, \bibinfo{author}{Lin,
  Z.}, \bibinfo{author}{Qin, T.}, \bibinfo{author}{Bian, J.},
  \bibinfo{author}{Liu, T.Y.}, \bibinfo{year}{2019}a.
\newblock \bibinfo{title}{Fully parameterized quantile function for
  distributional reinforcement learning}.
\newblock \bibinfo{journal}{Advances in neural information processing systems}
  \bibinfo{volume}{32}.
\bibitem[{Yang et~al.(2019b)Yang, Deng, Wang, Li, Lu, Che, Wei and
  Loparo}]{yang2019scalable}
\bibinfo{author}{Yang, S.}, \bibinfo{author}{Deng, B.}, \bibinfo{author}{Wang,
  J.}, \bibinfo{author}{Li, H.}, \bibinfo{author}{Lu, M.},
  \bibinfo{author}{Che, Y.}, \bibinfo{author}{Wei, X.},
  \bibinfo{author}{Loparo, K.A.}, \bibinfo{year}{2019}b.
\newblock \bibinfo{title}{Scalable digital neuromorphic architecture for
  large-scale biophysically meaningful neural network with multi-compartment
  neurons}.
\newblock \bibinfo{journal}{IEEE transactions on neural networks and learning
  systems} \bibinfo{volume}{31}, \bibinfo{pages}{148--162}.
\bibitem[{Zeng et~al.(2022)Zeng, Zhao, Zhao, Shen, Dong, Lu, Zhang, Sun, Liang,
  Zhao et~al.}]{zeng2022braincog}
\bibinfo{author}{Zeng, Y.}, \bibinfo{author}{Zhao, D.}, \bibinfo{author}{Zhao,
  F.}, \bibinfo{author}{Shen, G.}, \bibinfo{author}{Dong, Y.},
  \bibinfo{author}{Lu, E.}, \bibinfo{author}{Zhang, Q.}, \bibinfo{author}{Sun,
  Y.}, \bibinfo{author}{Liang, Q.}, \bibinfo{author}{Zhao, Y.}, et~al.,
  \bibinfo{year}{2022}.
\newblock \bibinfo{title}{Braincog: A spiking neural network based
  brain-inspired cognitive intelligence engine for brain-inspired ai and brain
  simulation}.
\newblock \bibinfo{journal}{arXiv preprint arXiv:2207.08533} .
\bibitem[{Zhang et~al.(2021a)Zhang, Cao, Zhang, Zhou and
  Feng}]{zhang2021distilling}
\bibinfo{author}{Zhang, L.}, \bibinfo{author}{Cao, J.}, \bibinfo{author}{Zhang,
  Y.}, \bibinfo{author}{Zhou, B.}, \bibinfo{author}{Feng, S.},
  \bibinfo{year}{2021}a.
\newblock \bibinfo{title}{Distilling neuron spike with high temperature in
  reinforcement learning agents}.
\newblock \bibinfo{journal}{arXiv preprint arXiv:2108.10078} .
\bibitem[{Zhang et~al.(2021b)Zhang, Wang, Wu, Belatreche, Amornpaisannon,
  Zhang, Miriyala, Qu, Chua, Carlson et~al.}]{zhang2021rectified}
\bibinfo{author}{Zhang, M.}, \bibinfo{author}{Wang, J.}, \bibinfo{author}{Wu,
  J.}, \bibinfo{author}{Belatreche, A.}, \bibinfo{author}{Amornpaisannon, B.},
  \bibinfo{author}{Zhang, Z.}, \bibinfo{author}{Miriyala, V.P.K.},
  \bibinfo{author}{Qu, H.}, \bibinfo{author}{Chua, Y.},
  \bibinfo{author}{Carlson, T.E.}, et~al., \bibinfo{year}{2021}b.
\newblock \bibinfo{title}{Rectified linear postsynaptic potential function for
  backpropagation in deep spiking neural networks}.
\newblock \bibinfo{journal}{IEEE transactions on neural networks and learning
  systems} \bibinfo{volume}{33}, \bibinfo{pages}{1947--1958}.
\bibitem[{Zhang et~al.(2024)Zhang, Yang, Ma, Wu, Li and Tan}]{zhang2024tc}
\bibinfo{author}{Zhang, S.}, \bibinfo{author}{Yang, Q.}, \bibinfo{author}{Ma,
  C.}, \bibinfo{author}{Wu, J.}, \bibinfo{author}{Li, H.},
  \bibinfo{author}{Tan, K.C.}, \bibinfo{year}{2024}.
\newblock \bibinfo{title}{Tc-lif: A two-compartment spiking neuron model for
  long-term sequential modelling}, in: \bibinfo{booktitle}{Proceedings of the
  AAAI Conference on Artificial Intelligence}, pp.
  \bibinfo{pages}{16838--16847}.
\bibitem[{Zhang et~al.(2022)Zhang, Chen, Zhang, Luo, Zhang, Qu and
  Yi}]{zhang2022minicolumn}
\bibinfo{author}{Zhang, Y.}, \bibinfo{author}{Chen, Y.},
  \bibinfo{author}{Zhang, J.}, \bibinfo{author}{Luo, X.},
  \bibinfo{author}{Zhang, M.}, \bibinfo{author}{Qu, H.}, \bibinfo{author}{Yi,
  Z.}, \bibinfo{year}{2022}.
\newblock \bibinfo{title}{Minicolumn-based episodic memory model with spiking
  neurons, dendrites and delays}.
\newblock \bibinfo{journal}{IEEE transactions on neural networks and learning
  systems} .
\bibitem[{Zhao et~al.(2018)Zhao, Zeng and Xu}]{zhao2018brain}
\bibinfo{author}{Zhao, F.}, \bibinfo{author}{Zeng, Y.}, \bibinfo{author}{Xu,
  B.}, \bibinfo{year}{2018}.
\newblock \bibinfo{title}{A brain-inspired decision-making spiking neural
  network and its application in unmanned aerial vehicle}.
\newblock \bibinfo{journal}{Frontiers in neurorobotics} \bibinfo{volume}{12},
  \bibinfo{pages}{56}.

\end{thebibliography}

%

\end{document}